\documentclass[runningheads]{llncs}
\usepackage{graphicx}

\usepackage{orcidlink}
\usepackage{tikz}
\usepackage{comment}
\usepackage{amsmath,amssymb} 
\usepackage{color}
\usepackage{ragged2e}
\usepackage{xcolor}         
\usepackage{graphicx}
\usepackage{subcaption}
\usepackage{placeins}
\usepackage[export]{adjustbox}
\usepackage{caption}

\usepackage{float}
\usepackage[utf8]{inputenc} 
\usepackage[T1]{fontenc}    
\usepackage{hyperref}       
\usepackage{url}            
\usepackage{booktabs}       
\usepackage{amsmath}
\usepackage{amsfonts}       
\usepackage{nicefrac}       
\usepackage{microtype}      

\usepackage[accsupp]{axessibility}  

\usepackage{etoolbox}
\newcommand{\repthanks}[1]{\textsuperscript{\ref{#1}}}
\makeatletter
\patchcmd{\maketitle}
  {\def\thanks}
  {\let\repthanks\repthanksunskip\def\thanks}
  {}{}
\patchcmd{\@maketitle}
  {\def\thanks}
  {\let\repthanks\@gobble\def\thanks}
  {}{}
\newcommand\repthanksunskip[1]{\unskip{}}
\makeatother

\begin{document}
\pagestyle{headings}
\mainmatter
\def\ECCVSubNumber{}  

\title{SKDCGN: Source-free Knowledge Distillation of Counterfactual Generative Networks using cGANs} 


\titlerunning{SKDCGN}
%


\author{Sameer Ambekar \orcidlink{0000-0002-8650-3180}\thanks{Equal contribution.\protect\label{contrib}} \and Matteo Tafuro \orcidlink{0000-0002-6167-2156}\repthanks{contrib} \and
Ankit Ankit \orcidlink{0000-0002-9399-9209}\repthanks{contrib}\and \\Diego van der Mast\index{van der Mast, Diego} \orcidlink{0000-0002-0001-3069}\repthanks{contrib} \and Mark Alence \orcidlink{0000-0002-6622-5822}\repthanks{contrib} \and Christos  Athanasiadis \orcidlink{0000-0003-4376-9066}}
\authorrunning{S. Ambekar et al.}
\institute{University of Amsterdam, Amsterdam, the Netherlands. \\
\email{ambekarsameer@gmail.com,
tafuromatteo00@gmail.com, ankitnitt1721@gmail.com,
diego.vandermast@student.uva.nl,
mark.alence@gmail.com, c.athanasiadis@uva.nl }}
\maketitle

\begin{abstract}
\justifying{With the usage of appropriate inductive biases, Counterfactual Generative Networks (CGNs) can generate novel images from random combinations of shape, texture, and background manifolds. These images can be utilized to train an invariant classifier, avoiding the wide spread problem of deep architectures learning spurious correlations rather than meaningful ones. As a consequence, out-of-domain robustness is improved. However, the CGN architecture comprises multiple over parameterized networks, namely BigGAN and U2-Net. Training these networks requires appropriate background knowledge and extensive computation. Since one does not always have access to the precise training details, nor do they always possess the necessary knowledge of counterfactuals, our work addresses the following question: Can we use the knowledge embedded in pre-trained CGNs to train a lower-capacity model, assuming black-box access (i.e., only access to the pretrained CGN model) to the components of the architecture? In this direction, we propose a novel work named SKDCGN that attempts knowledge transfer using Knowledge Distillation (KD). In our proposed architecture, each independent mechanism (shape, texture, background) is represented by a student 'TinyGAN' that learns from the pretrained teacher 'BigGAN'. We demonstrate the efficacy of the proposed method using state-of-the-art datasets such as ImageNet, and MNIST by using KD and appropriate loss functions.
Moreover, as an additional contribution, our paper conducts a thorough study on the composition mechanism of the CGNs, to gain a better understanding of how each mechanism influences the classification accuracy of an invariant classifier. Code available at: \url{https://github.com/ambekarsameer96/SKDCGN}}
\end{abstract}


\section{Introduction}
\label{sec:intro}
Deep neural networks are prone to learning simple functions that fail to capture intricacies of data in higher-dimensional manifolds \cite{DBLP:journals/corr/abs-2110-02424}, which causes networks to struggle in generalizing to unseen data. In addition to spectral bias \cite{DBLP:journals/corr/abs-2110-02424} and shortcut learning, which are properties inherent to neural networks \cite{DBLP:journals/corr/abs-2004-07780}, spurious learned correlations are also caused by biased datasets.
To this end, Counterfactual Generative Networks (CGNs), proposed by Sauer and Geiger \cite{DBLP:journals/corr/abs-2101-06046}, have been shown to generate novel images that mitigate this effect. The authors expose the causal structure of image generation and split it into three Independent Mechanisms (IMs) (object shape, texture, and background), to generate synthetic and \textit{counterfactual} images whereon an invariant classifier ensemble can be trained. 

The CGN architecture comprises multiple over-parameterized networks, namely BigGANs \cite{brock2019large} and U2-Nets \cite{DBLP:journals/corr/abs-2005-09007}, and its training procedure generally requires appropriate domain-specific expertise. Moreover, one does not always have access to the precise training details, nor do they necessarily possess the required knowledge of counterfactuals. Motivated by these observations, we propose \textit{Source-free Knowledge Distillation of Counterfactual Generative Networks} (SKDCGN), which aims to use the knowledge embedded in a pre-trained CGN to train a lower capacity model, assuming black-box access (i.e., only inputs and outputs) to the components of the source model. More specifically, we harness the idea of Knowledge Distillation (KD) \cite{DBLP:journals/corr/abs-2106-05237} to train a network comprising three (small) generative models, i.e. TinyGANs \cite{DBLP:journals/corr/abs-2009-13829}, each being responsible for a single independent mechanism. SKDCGN carries both practical and theoretical implications, and it is intended to:
\begin{enumerate}
    \item Obtain a lightweight version of the CGN, reducing its computational cost and memory footprint. This is meant to (i) ease the generation of counterfactual datasets and hence encourage the development of robust and invariant classifiers, as well as (ii) potentially allowing the deployment of the model on  resource-constrained devices.
    \item Explore whether we can \textit{learn} from a fully trained CGN and distill it to a less parameterized network, assuming that we do not have access to the training process of the model.
\end{enumerate}

Along the lines of the original paper, we demonstrate the ability of our model to generate counterfactual images on ImageNet-1k \cite{5206848} and Double-Colored MNIST \cite{DBLP:journals/corr/abs-2101-06046}. Furthermore, we compare our outputs to \cite{DBLP:journals/corr/abs-2101-06046} and a simple baseline in terms of out-of-distribution robustness on the original classification task. As an additional contribution, we conduct a study on the shape IM of the CGN.

The paper is organized as follows: firstly, we present a brief literature survey in Section \ref{sec:related-work}; next in Section \ref{sec:approach} the SKDCGN is dissected; Section \ref{sec:exps-results} presents the experimental setup and the empirical results, which are finally discussed in Section \ref{sec:conclusion}.

\section{Related work}
\label{sec:related-work}
This section introduces the fundamental concepts and the related works that we use as a base for our SKDCGN.

\subsubsection{Counterfactual Generative Networks. }
The main idea of CGNs \cite{DBLP:journals/corr/abs-2101-06046} has already been introduced in Section \ref{sec:intro}. Nonetheless, to aid the understanding of our method to readers that are not familiar with the CGN architecture, we summarize its salient components in this paragraph and also provide the network diagram in Appendix \ref{app:cgn-architecture}, Figure \ref{fig:cgn-diagram}. The CGN consists of 4 backbones: (i) the part of the network responsible for the shape mechanism, those responsible for (ii) texture and (iii) background, and a (iv) composition mechanism that combines the previous three using a deterministic function. Given a noise vector $\mathbf{u}$ (sampled from a spherical Gaussian) and a label $y$ (drawn uniformly from the set of possible labels y) as input, (i) the shape is obtained from a BigGAN-deep-256 \cite{brock2019large}, whose output is subsequently passed through a U2-Net \cite{DBLP:journals/corr/abs-2005-09007} to obtain a binary mask of the object shape. The (ii) texture and (iii) background are obtained similarly, but the BigGAN's output does not require to be segmented by the U2-Net. Finally, the (iv) composition mechanism outputs the final counterfactual image $\mathbf{x}_{gen}$ using the following analytical function:
\begin{equation}
\label{eq:composition}
    \mathbf{x}_{g e n}=C(\mathbf{m}, \mathbf{f}, \mathbf{b})=\mathbf{m} \odot \mathbf{f}+(1-\mathbf{m}) \odot \mathbf{b},
\end{equation}
where $\mathbf{m}$ is the shape mask, $\mathbf{f}$ is the foreground (or texture), $\mathbf{b}$ is the background and  $\odot$ denotes element-wise multiplication.

More recently, \cite{khorram2022cycleconsistent} devises an approach that learns a latent transformation that generates visual CFs automatically by steering in the latent space of generative models. Additionally, \cite{DBLP:journals/corr/abs-2109-14274} uses a deep model inversion approach that provides counterfactual explanations by examining the area of an image.

\subsubsection{Knowledge Distillation. } \cite{44873} firstly proposed to transfer the knowledge of a pre-trained cumbersome network (referred to as the \textit{teacher}) to a smaller model (the \textit{student}). This is possible because networks frequently learn low-frequency functions among other things, indicating that the learning capacity of the big network is not being utilized fully \cite{DBLP:journals/corr/abs-2110-02424} \cite{DBLP:journals/corr/abs-2004-07780}. Traditional KD approaches (often referred to as \textit{black-box}) simply use the outputs of the large deep model as
the teacher knowledge, but other variants have made use of activation, neurons or features of intermediate layers as the knowledge to guide the learning process \cite{kdref1,kdref2}. Existing methods like \cite{DBLP:journals/corr/abs-2009-13829} are also making use of Knowledge distillation for the task of image generation. Our work is similar to this, however, they transfer the knowledge of BigGAN trained on ImageNet dataset to a TinyGAN. In contrast, in our work, we transfer not just the knowledge of image generation but also the task of counterfactual generation from a BigGAN to a TinyGAN. 

\subsubsection{Distilling GANs using KD. } Given its high effectiveness for model compression, KD has been widely used in different fields, including visual recognition and classification, speech recognition, natural language processing (NLP), and recommendation systems \cite{kd-survey}. However, it is less studied for image generation. \cite{DBLP:journals/corr/abs-1902-00159} firstly applied KD to GANs. However, our project differs from theirs as they use \textit{unconditional} image generation, less general (DCGAN \cite{dcgan}) architectures and they do not assume a black-box generator. Our setting is much more similar to that of \cite{DBLP:journals/corr/abs-2009-13829}, where a BigGAN is distilled to a network with 16$\times$ fewer parameters, assuming no access to the teacher's training procedure or parameters. Considering its competitive performance, we use the proposed architecture (TinyGAN) as the student model and use a modified version of their loss function (further details in Section \ref{sec:method-training}) to optimize our network.


\textbf{Source-free}: We term our method as Source-free since we do not have access to the source data, source training details, procedure, and any knowledge about the counterfactuals, etc, but only have access to trained source models. This method is similar to methods such as \cite{yang2021generalized} \cite{ding2022source}. With large diffusion models like Imagen \cite{saharia2022photorealistic} and DALL·E 2 \cite{https://doi.org/10.48550/arxiv.2204.13807} where the training process is usually extremely expensive in terms of computation, lack precise details about training them and often not reproducible by academic groups, we often have access to pretrained models. These can be used to transfer knowledge to a smaller network, and perform the same task with model of lower capacity.













\section{Approach}
\label{sec:approach}

This section dives into the details of the SKDCGN architecture, focusing on the training and inference phases separately for ImageNet-1k and MNIST. In addition, we discuss the loss functions that were employed for Knowledge Distillation.

\subsection{SKDCGN}
\begin{figure}[t]
  \includegraphics[width=\linewidth]{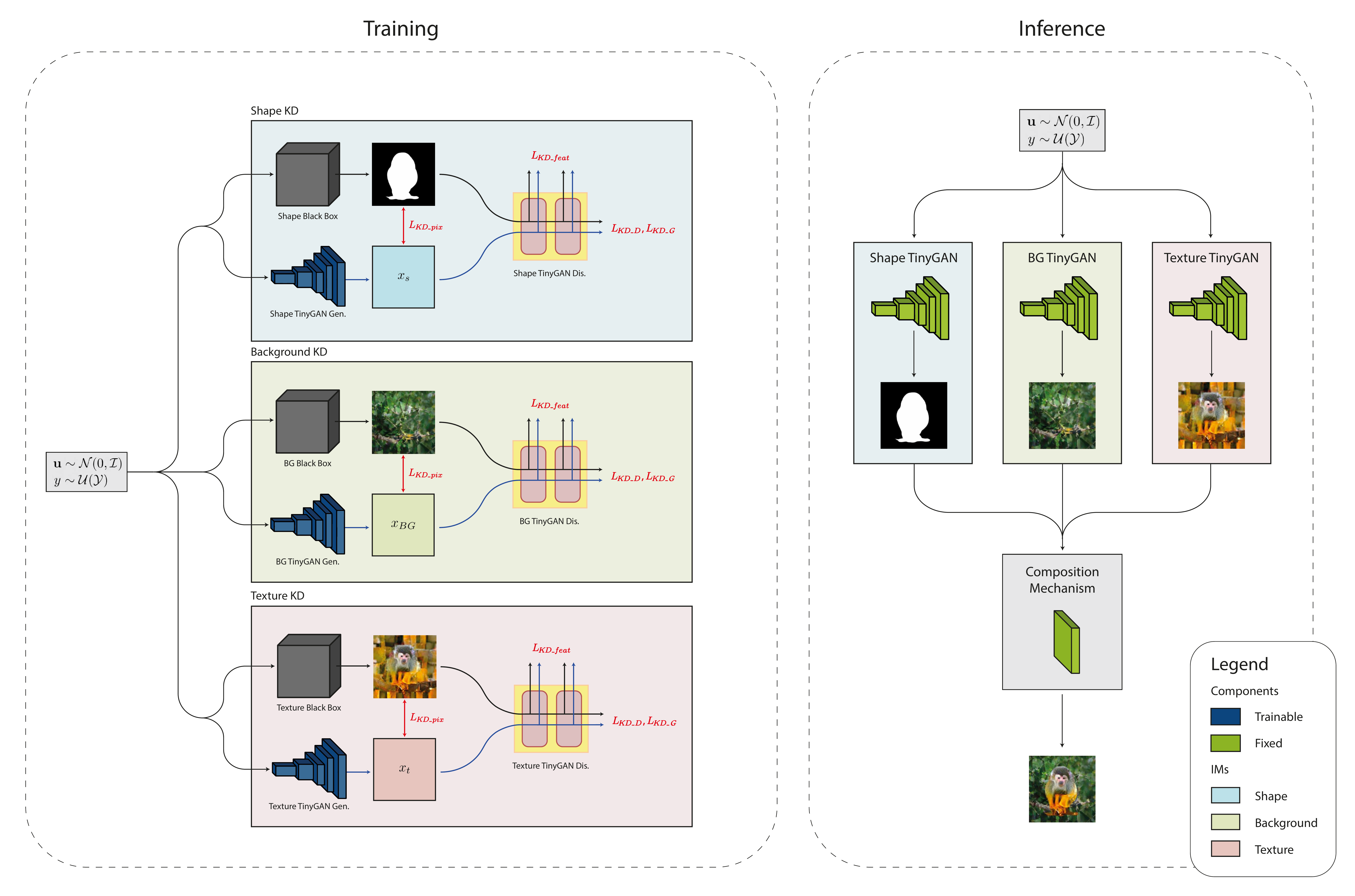}
  \caption{\textit{Architecture of the SKDCGN.} During training, each independent mechanism serves as a black-box teacher model to train a corresponding student model. During inference, the outputs of the three trained TinyGANs are combined using a Composition Mechanism that returns the final counterfactual image.}
  \label{fig:arch_diagram}
\end{figure}

Although transferring the knowledge of an entire CGN into a single generative model could drastically reduce the number of parameters, this strategy would compromise the whole purpose of CGNs, i.e. disentangling the three mechanisms and having control over each of them. Therefore, we opt to train a generative model for each individual component. As shown in the architecture diagram (Figure \ref{fig:arch_diagram}), we treat each IM backbone as a black-box teacher and aim to mimic its output by training a corresponding TinyGAN student. Note that this implies that in the case of the shape mechanism, a single generative model learns to mimic both the BigGAN and the U2-Net. We believe a TinyGAN should be capable of learning binary masks directly, removing the need for the U2-Net and reducing the model size even further. During inference, the outputs of the three students are combined into a final counterfactual image using the composition function defined in Equation \ref{eq:composition}.

\subsubsection{Training: Distilling the knowledge of IMs. }
\label{sec:method-training}

To train SKDCGN, we utilize each IM backbone from the CGN architecture as a black-box teacher for the student network, as visualized in the training section of Figure \ref{fig:arch_diagram} (the backbones are BigGAN + U2-Net for \textit{shape}, BigGAN for \textit{texture}, and BigGAN for \textit{background}). As introduced in the \hyperref[sec:related-work]{Related work} section, \cite{DBLP:journals/corr/abs-2009-13829} proposed an effective KD framework for compressing BigGANs. As the IMs in CGNs rely on BigGANs, we utilize their proposed student architecture. For completeness, the details of the student architecture are reported in 
Appendix \ref{app:tinygan-architecture}, Figure \ref{fig:tinygan-generator}.

We base our training objective on the loss function proposed by \cite{DBLP:journals/corr/abs-2009-13829}. Our full objective comprises multiple terms:
(i) a pixel-wise distillation loss, (ii) an adversarial distillation loss, (iii) a feature-level distillation loss, and (iv) KL Divergence. In addition to introducing KL Divergence, we deviate from the original TinyGAN training objective by omitting the term that allows the model to learn from real images of the ImageNet dataset. This would inevitably compromise the quality of the generated counterfactuals. KL Divergence leads to entropy minimization between the teacher and student, which is why we propose its usage.

The individual loss terms are dissected below as from \cite{DBLP:journals/corr/abs-2009-13829}:
\begin{enumerate}
    \item \textit{Pixel-wise Distillation Loss}: To imitate the functionality of BigGAN for scaling generation to high-resolution, high-fidelity images, we minimize the pixel-level distance (L1) between the images generated by BigGAN and TinyGAN given the same input:
    \begin{equation}
    \mathcal{L}_{\text{KD\_pix}} = \mathbb{E}_{z \sim p(z), y \sim q(y)}[\|T(z,y) - S(z,y) \|_{1}]
    \label{pixelwise_loss}
    \end{equation}
    where $T$ represents the Teacher network, $S$ represents the Student network, $z$ is a latent variable drawn from the truncated normal distribution $p(z)$, and $y$ is the class label sampled from some categorical distribution $q(y)$.

    \item \textit{Adversarial Distillation Loss}: To promote sharper outputs, an adversarial loss is incorporated to make the outputs of $S$ indistinguishable from those of $T$. It includes a loss for the generator (Eq. \ref{eq:loss-adv-gen}) and one for the discriminator (Eq. \ref{eq:loss-adv-dis}):
    \begin{align}
    \mathcal{L}_{\text{KD\_G}} =& - \mathbb{E}_{z, y}[D(S(z,y), y)] \label{eq:loss-adv-gen}\\
    \mathcal{L}_{\text{KD\_D}} =& - \mathbb{E}_{z, y}\left[max(0, 1 - D(T(z,y), y)) + max(0, 1 - D(S(z,y), y))\right] \label{eq:loss-adv-dis},
    \end{align}
    where $z$ is the noise vector, $y$ is the class label, $T(z,y)$ is the image generated by the Teacher $T$, while $G$ and $D$ are -- respectively -- the generator and discriminator of the Student $S$.
    
    \item \textit{Feature Level Distillation Loss}: To further overcome the blurriness in the images produced by the Student network, the training objective also includes a feature-level distillation loss. More specifically, we take the features computed at each convolutional layer in the Teacher discriminator, and with a loss function stimulate $S$ to generate images similar to $T$:
    \begin{equation}
    \mathcal{L}_{\text{KD\_feat}} = \mathbb{E}_{z, y}\left[\sum _{i} \alpha_{i}\left\|D_{i}(T(z,y),y) - D_{i}(S(z,y), y) \right\|_{1}\right]
    \label{feature_loss}
    \end{equation}
    where $D_{i}$ represents the feature vector extracted from the $i^{th}$ layer of the discriminator and the corresponding weights are given by $\alpha_{i}$.

    \item \textit{KL Divergence}: L1 alone cannot reduce the entropy between the teacher and target. To improve the proposed method, we use KL Divergence in a similar fashion to \cite{asano2021extrapolating} for the task of knowledge distillation between real images drawn from source $P(x)$ and target images $Q(x)$. The 
    \begin{equation}
    \mathcal D_{\mathrm{KL}}(P \| Q)=\sum_{x \in \mathcal{X}} P(x) \log \left(\frac{P(x)}{Q(x)}\right)
    \label{feature_loss_kl}
    \end{equation}
    
    \begin{equation}
    \mathcal{L}_{\text{KL}} = \sum_{x \in X}-p_{x}^{t} \log p_{x}^{s}+p_{x}^{t} \log p_{x}^{t}
    \label{eq:kl-loss}
    \end{equation}
    where $x$ is the class label and $p$ contains the output softmax probabilities of the Generator $G$ divided by the temperature $t$. 
    
\end{enumerate}
To sum up, the student's generator ($G$) and discriminator ($D$) are respectively optimized using the following objectives:
\begin{align}
    \mathcal{L}_{\text{G}} = & \mathcal{L}_{\text{KD\_feat}} +
    \lambda_1 \mathcal{L}_{\text{KD\_pix}} + \lambda_2\mathcal{L}_{\text{KD}\_G}
    \,(\;+\;\mathcal{L}_{\text{KL}}\,)\\
    \mathcal{L}_{\text{D}} = & \mathcal{L}_{\text{KD\_D}}
\end{align}
where $\lambda_1$ and $\lambda_2$ are the regularization terms mentioned in \cite{DBLP:journals/corr/abs-2009-13829}, and the KL divergence term ($\mathcal{L}_{\text{KL}}$) is only used in the enhanced version of SKDCGN. 

Implementing the SKDCGN architecture requires training a TinyGAN for each Independent Mechanism of the CGN (see Fig. \ref{fig:arch_diagram}). The KD training procedure, however, requires training data. Hence prior to training, 1000 images per class (totalling 1 million samples) are generated using the IM backbones
extracted from the pre-trained CGN (as provided by Sauer and Geiger \cite{DBLP:journals/corr/abs-2101-06046}).

Finally, note that the original CGN architecture (illustrated in Appendix \ref{app:cgn-architecture}, Figure \ref{fig:cgn-diagram}) comprises another BigGAN trained on ImageNet-1k. It is unrelated to the three Independent Mechanisms and provides primary training supervision via reconstruction loss. We discard this component of the architecture for two main reasons: we do not have a dataset of counterfactuals whereon a GAN can be trained; we argue that this additional knowledge is already embedded in the backbones of a pre-trained CGN.

\subsubsection{Inference: generating counterfactuals. }
Once the three student networks are trained, their outputs are combined during inference akin to \cite{DBLP:journals/corr/abs-2101-06046} using the analytical function of Equation \ref{eq:composition}. Since the composition function is deterministic, we devise inference as a separate task to training. 

\section{Experiments and results}
\label{sec:exps-results}
This section defines our experimental setup, then proceeds to present the results. First, we test SKDCGN -- as defined in the \hyperref[sec:approach]{Approach} section -- on both ImageNet-1k and MNIST (Section \ref{sec:exps-skdcgn}), and based on the observed findings we make some changes to the proposed architecture to improve the quality of the results (Section \ref{sec:exps-improvement}). Due to computational constraints we test these improvements on a smaller dataset, namely the double-colored variant of MNIST \cite{726791}. Finally, as an additional contribution, we conduct a thorough study on the composition mechanism, to gain a better understanding of how each mechanism influences the classification accuracy of an invariant classifier. We present the results of such a study in Section \ref{sec:exps-comp-mechanism}.


\subsection{Datasets}
\paragraph{ImageNet-1k.} The ImageNet-1k ILSVRC dataset \cite{5206848} contains 1,000 classes, with each class consisting of 1.2 million training images, 50,000 validation and 100,000 test images. Images were resized to $256\times256$ to maintain consistent experiments and to allow direct comparisons with the original results of \cite{DBLP:journals/corr/abs-2101-06046}.

\paragraph{Double-colored MNIST.} We use the \textit{double-colored} MNIST dataset proposed by Sauer and Geiger in the original CGN paper \cite{DBLP:journals/corr/abs-2101-06046}.  This is a variant of the MNIST dataset where both the digits and the background are independently colored. It consists of 60,000 $28\times28$ images of the 10 digits, along with a test set of 10,000 images.


\subsection{Baseline Model: CGN with generator replaced by TinyGAN generator}
The SKDCGN is compared with a modified version of the original CGN architecture, where each BigGAN has been replaced by the generator model of a TinyGAN. Training this baseline using the procedure described by \cite{DBLP:journals/corr/abs-2009-13829}, omitting KD, allows for rigorous comparisons that emphasize the effectiveness of the knowledge distillation process. Further training details are provided in Appendix \ref{app:baseline-training}.

\subsection{Results of SKDCGN}
\label{sec:exps-skdcgn}

\begin{figure}[t]
   \begin{subfigure}{\textwidth}
   \centering
   \hspace{6mm} \textit{ImageNet-1k} \hspace{36mm} \textit{Double-colored MNIST}\\
      \includegraphics[width=0.48\linewidth]{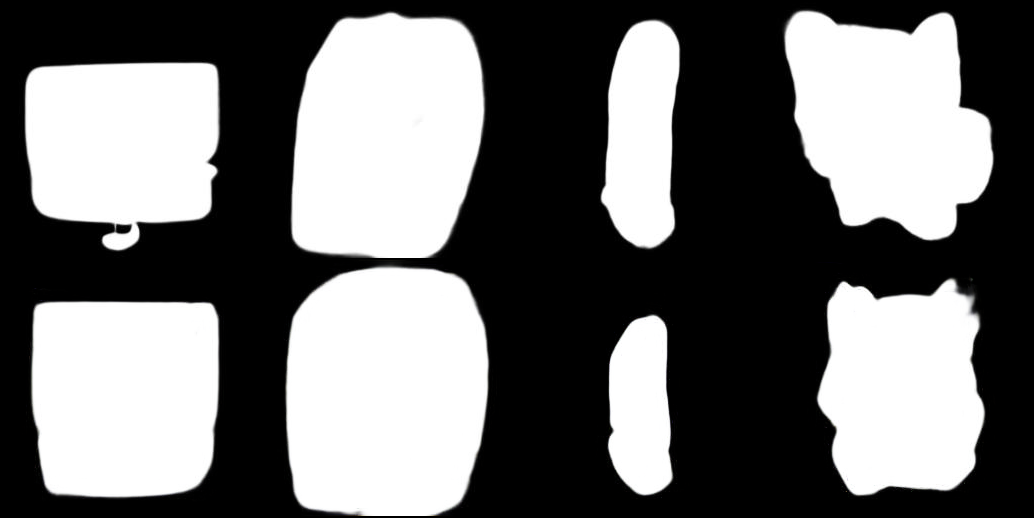}
      \hfill
      \includegraphics[width=0.48\linewidth]{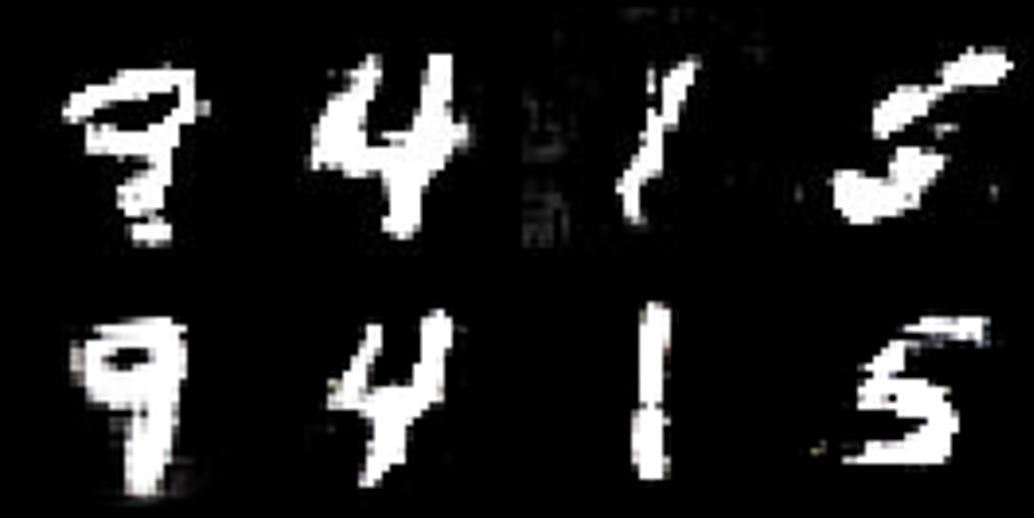}
      \caption{\textit{Shape} mechanism.}
      \label{fig:shape_results}
    \end{subfigure}
    \\
  \begin{subfigure}{\textwidth}
  \centering
  \hspace{6mm} \textit{ImageNet-1k} \hspace{36mm} \textit{Double-colored MNIST}\\
      \includegraphics[width=0.48\linewidth]{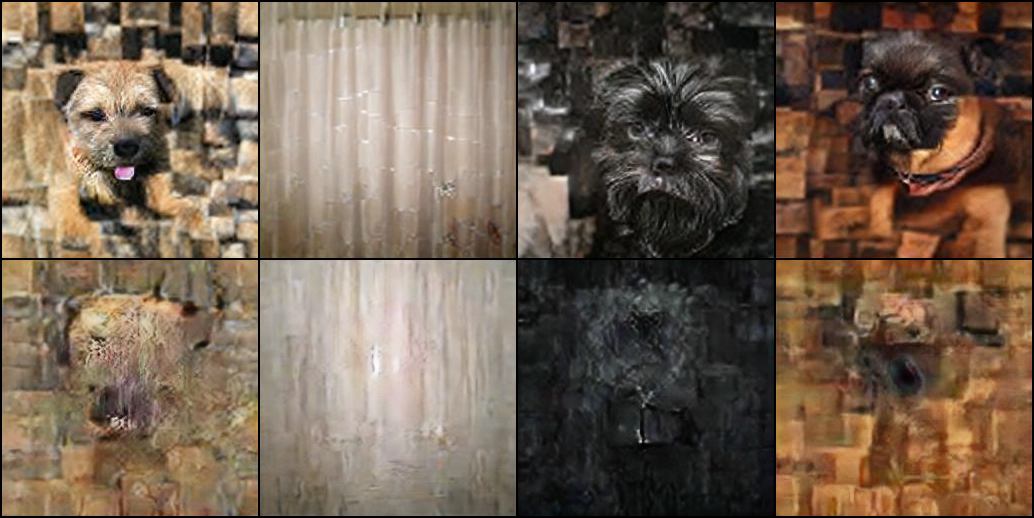}
      \hfill
      \includegraphics[width=0.48\linewidth]{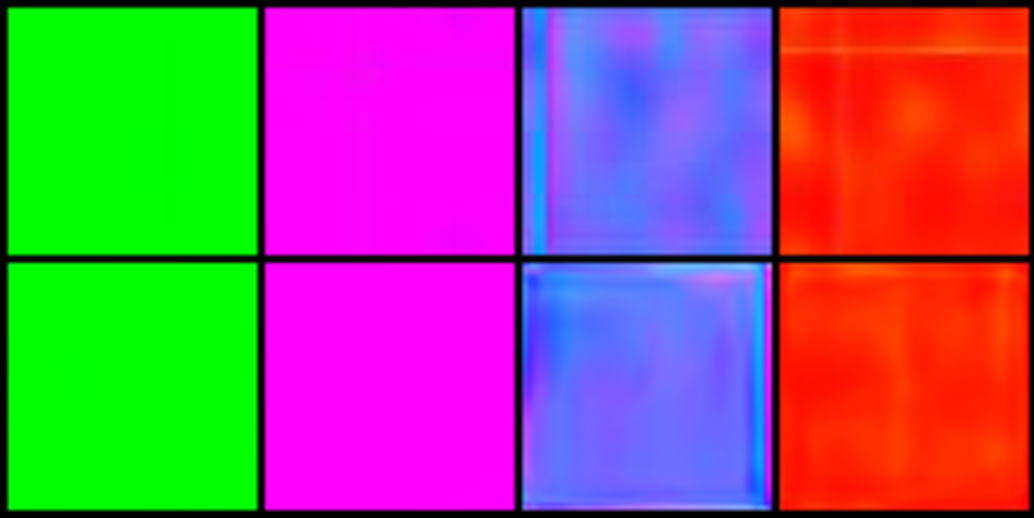}
      \caption{\textit{Texture} mechanism.}
      \label{fig:fg_results}
  \end{subfigure}
  \\
    \begin{subfigure}{\textwidth}
    \centering
      \hspace{6mm} \textit{ImageNet-1k} \hspace{36mm} \textit{Double-colored MNIST}\\
      \includegraphics[width=0.48\linewidth]{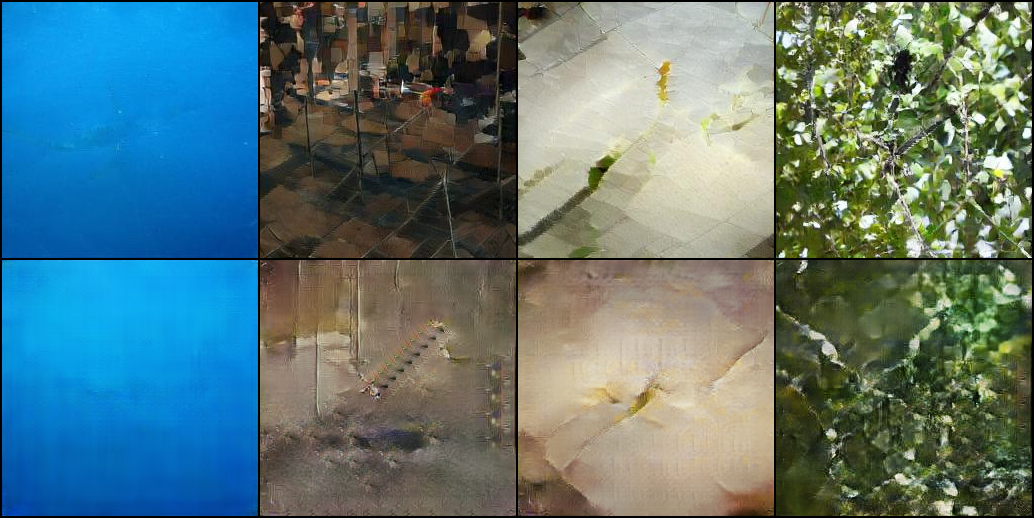}
      \hfill
      \includegraphics[width=0.48\linewidth]{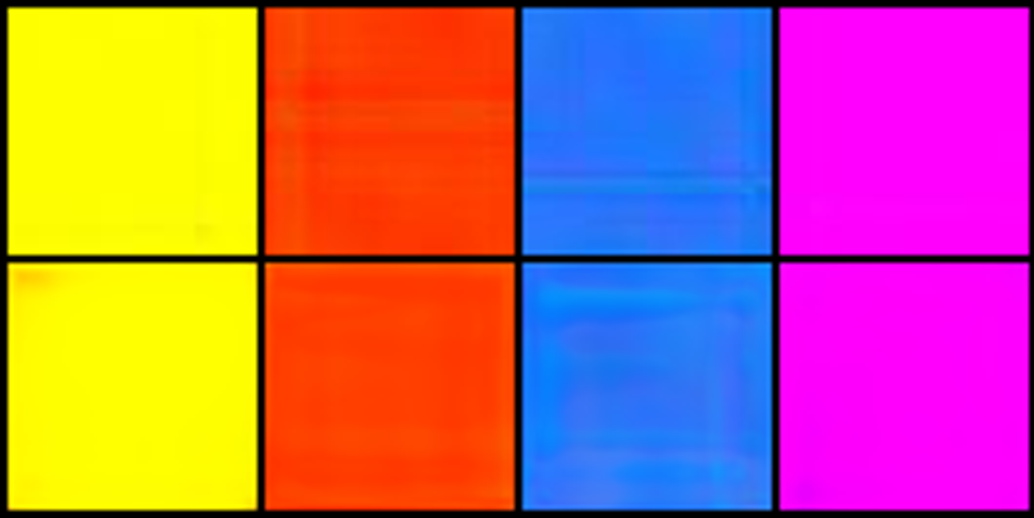}
      \caption{\textit{Background} mechanism.}
      \label{fig:bg_results}
  \end{subfigure}
  \caption{A comparison of images (on both ImageNet-1k and double-colored MNIST) generated by the CGN backbones and those generated by the corresponding SKDCGN's TinyGAN (given the same input), for each independent mechanism.}
  \label{fig:im-results_t_b}
\end{figure}

The proposed model was firstly trained and tested on ImageNet-1k. To further validate our method, we repeated the training procedure on MNIST. 

The qualitative results are collected in Figure \ref{fig:im-results_t_b} and demonstrate that TinyGANs can closely approximate the output of each IM. While this is true for both datasets, the effectiveness of our method is especially visible in the case of MNIST. It is likely the case that the reduced capacity of the TinyGANs (compared to the original CGN backbones) is sufficient to decently model the underlying data distribution. ImageNet-1k, on the other hand, reveals more apparent (though still acceptable) discrepancies between the images, especially for the \textit{texture} IM.

However, careful and extensive experiments revealed that the three TinyGANs could not generalize when random noise was given to the generator, i.e., they could not produce results beyond the test set. This might be due to a number of reasons. First, the compromised generalization capabilities of each IM's TinyGAN could be caused by their reduced network capacity. Furthermore, each TinyGAN was trained on all 1000 classes of ImageNet-1K, as opposed to Chang and Lu's choice of limiting the training data to the 398 animal labels \cite{DBLP:journals/corr/abs-2009-13829}. Finally, we generate the test samples using the test noise instead of random noise, since we hypothesize that the student networks only learn the manifolds that the teacher networks have been trained on. Additional experiments are required to analyze whether samples generated using random noise are found along the same manifold; unfortunately, we were hindered by the limited time frame allocated for this project, hence we leave this question open for future works.


\begin{figure}[t!]
\centering
  \begin{subfigure}{\textwidth}
      \includegraphics[width=\linewidth]{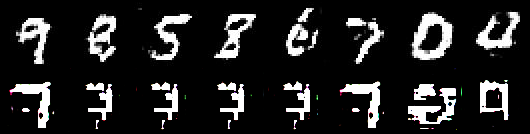}
      \caption{\textit{Shape} mechanism.}
      \label{fig:mnist_mask_kl_div_fg}
  \end{subfigure}
  \\
  \begin{subfigure}{\textwidth}
      \includegraphics[width=\linewidth]{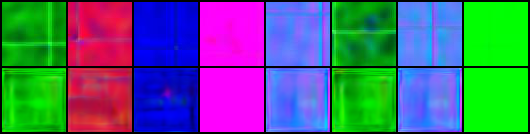}
      \caption{\textit{Texture} mechanism.}
      \label{fig:mnist_mask_kl_div_bg}
  \end{subfigure}
  \\
    \begin{subfigure}{\textwidth}
      \includegraphics[width=\linewidth]{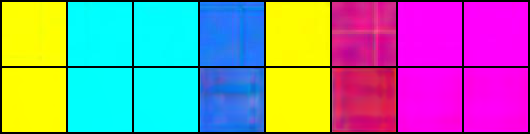}
      \caption{\textit{Background} mechanism.}
      \label{fig:mnist__mask_kl_div_mask}
  \end{subfigure}
  \caption{A comparison of double-colored MNIST images generated by the CGN backbones and those generated by the corresponding SKDCGN's TinyGAN (given the same input) for each IM. Here, SKDCGN was tuned such that KL divergence is minimized between the teacher and student networks, and the L1 loss is multiplied with the activation of every layer.}
  \label{fig:mnist_kl_div}
\end{figure}

\begin{figure}[t!]
  \centering
  \begin{subfigure}{0.47\textwidth}
      \includegraphics[width=\linewidth]{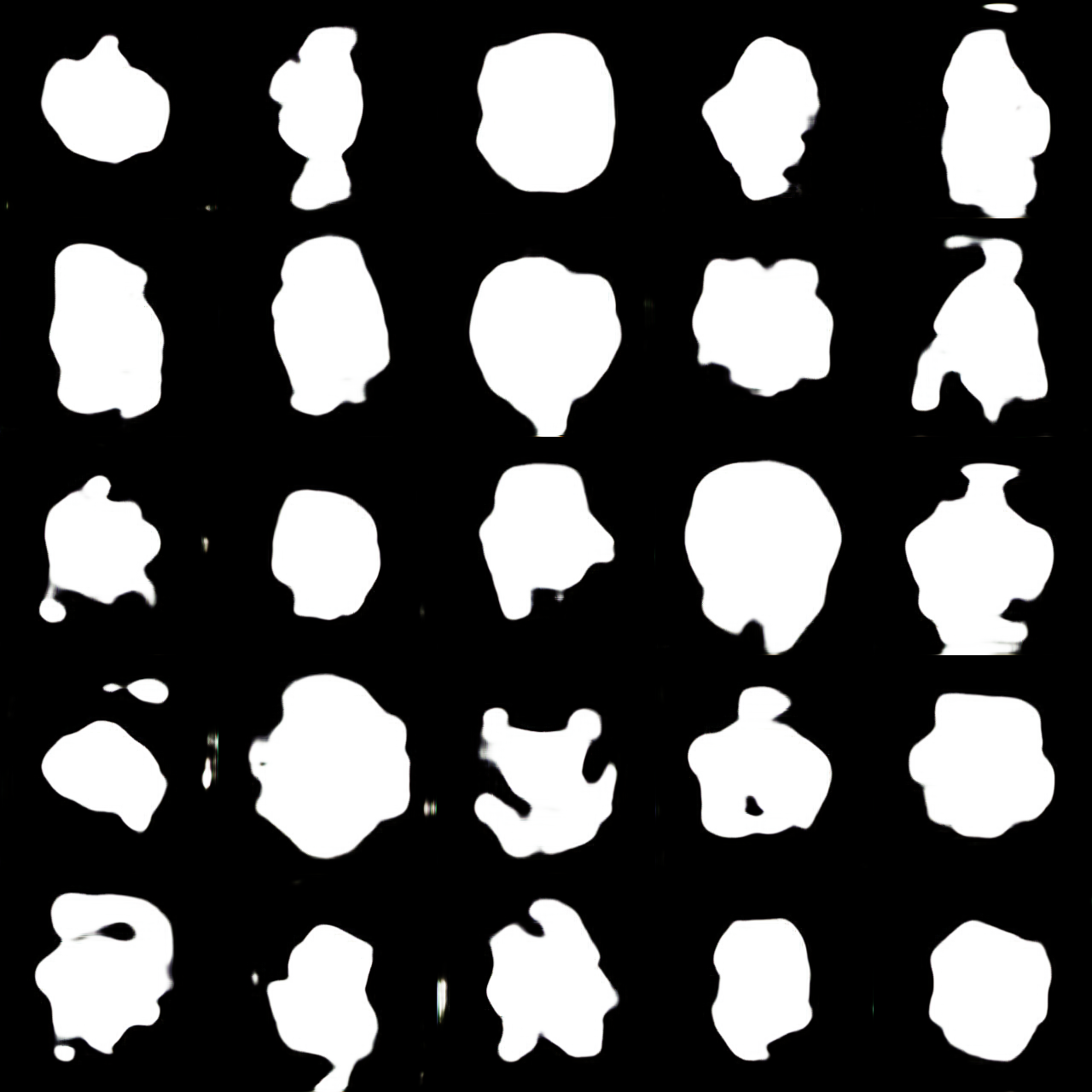}
      \caption{}
    \end{subfigure}
    \hfill
    \begin{subfigure}{0.47\textwidth}
      \includegraphics[width=\linewidth]{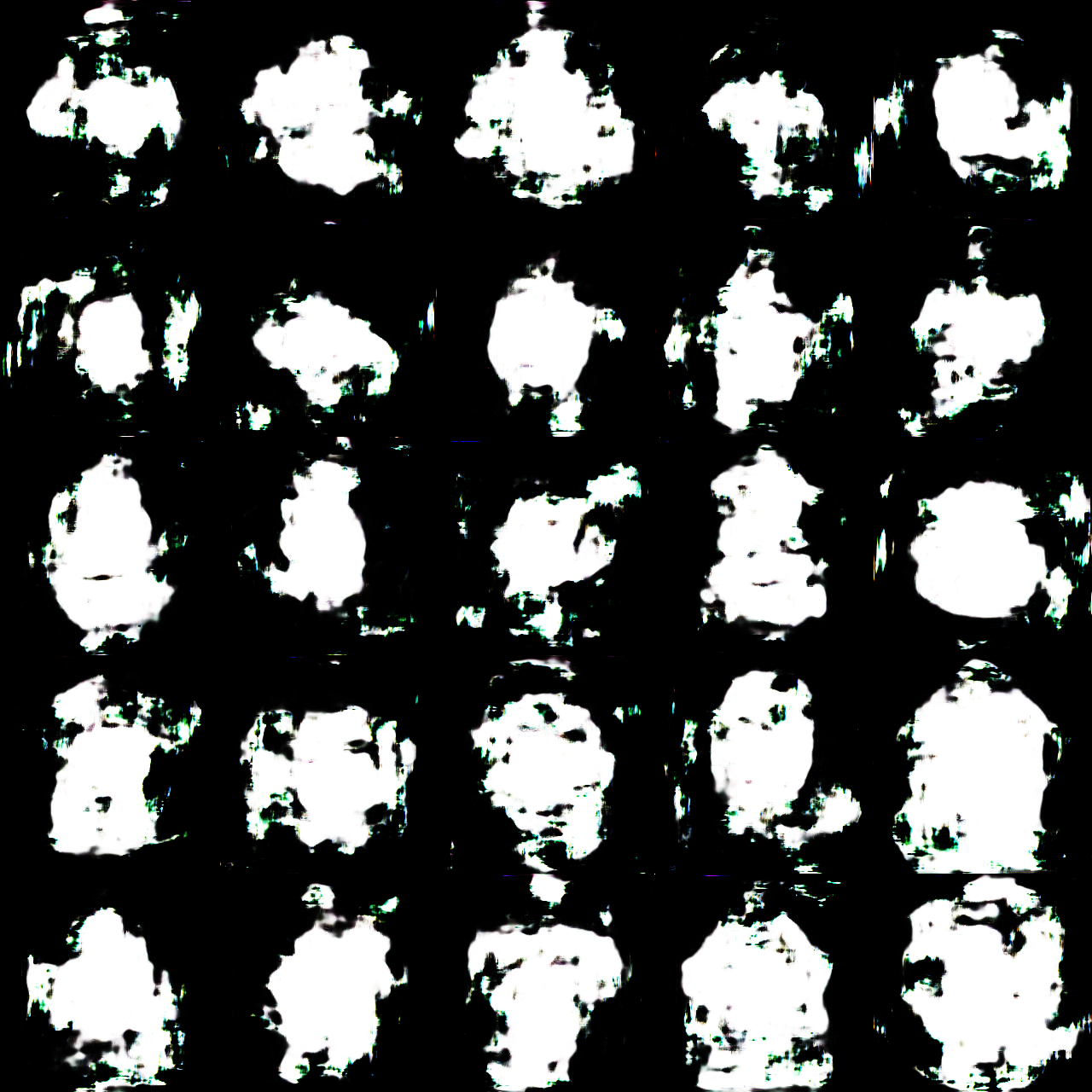}
      \caption{}
    \end{subfigure}
  \caption{(a) Shape masks obtained after the \textit{first} epoch of SKDCGN training on ImageNet-1k, using KL divergence. (b) Shape masks obtained after the 23$^{\text{rd}}$ epoch of SKDCGN training on ImageNet-1k, \textit{without} KL divergence. Evidently, KL enhances the quality of the masks from the first epoch, whereas its absence compromises the results even at a later stage of training.}
  \label{fig:Imagenet_mask_kl_div}
\end{figure}

\subsection{Improving the SKDCGN model}
\label{sec:exps-improvement}

The results presented in the previous section reveal that the outputs are noisy and ambiguous in nature when knowledge distillation is performed using the pre-trained models provided by Sauer and Geiger \cite{DBLP:journals/corr/abs-2101-06046} (note the artifacts in the SKDCGN's outputs of Fig. \ref{fig:im-results_t_b}, especially those trained on ImageNet-1k). This statement was supported by an interesting yet unexpected result of the study on the composition mechanism (refer to Section \ref{sec:exps-comp-mechanism}): it was observed that modifying Equation \ref{eq:composition} such that the shape mask $\mathbf{m}$ is multiplied with a weight factor of 0.75 (i.e., setting the transparency of the shape mask to 75\%), yielded an accuracy increase of the CGN's invariant classifier. The findings of this experiment -- conducted on the double-colored MNIST dataset -- suggest that the mask component is noisy in nature, leading to ambiguities in the decision boundaries during the classification of several digits. 

In light of this new hypothesis, we attempt to use the \textit{Kullback–Leibler} (KL) divergence to improve the visual quality of the outputs\footnote{It is noteworthy that other techniques were tested in the attempt to improve the visual quality of the results. Although they did not prove to be as beneficial, they are described in Appendix \ref{sec:improve_skdcgn}.}. Since KL leads to entropy minimization between the teacher and student networks, we deem such a technique adequate for the task at hand. Moreover, the choice of using KL was encouraged by the work of Asano and Saeed \cite{asano2021extrapolating}, which proved the suitability of the measure in this context. Concretely, the KL Divergence loss (as defined in Eq. \ref{eq:kl-loss}) was included in the overall generator loss $\mathcal{L}_{\text{G}}$ as seen in Equation \ref{eq:loss-adv-gen}. 


First, the modified SKDCGN was tested on the double-colored MNIST dataset. As depicted in Figure \ref{fig:mnist_kl_div}, the introduction of KL divergence improves SKDCGN's visual fidelity of both \textit{background} and \textit{texture} IMs, while the quality of the \textit{shape} masks seems to diminish after a few epochs. Contrarily, this approach appeared to be beneficial for the shape mechanism too, in the context of ImageNet-1k. The shape masks resulted more natural and consistent since the first epoch, whereas the absence of KL yielded noisy masks even at a later stage of training (refer to Figure \ref{fig:Imagenet_mask_kl_div}).

\subsection{Additional results: study of the shape IM}
\label{sec:exps-comp-mechanism}

\begin{table}[t]
\centering
\begin{tabular}{lrrr}
\toprule
& \;\;Noise & \;\;Rotation & \;\;Transparency\\
\midrule
Train Accuracy & $99.9$ & $99.1$ & $94.7$ \\
Test Accuracy & $14.96$ & $13.51$ & $\mathbf{58.86}$  \\
\bottomrule\\
\end{tabular}
\caption{Results of the invariant classifier for the analysis of the shape IM. The classifier has been trained to predict whether images are CGN-generated or real. The training examples contain counterfactuals whose shape mechanism has been tuned with one of the three transformations indicated in the table (noise, rotation, transparency -- refer to Sec.\ref{sec:exps-comp-mechanism} for further details).}
\label{tab:shape_exp_results}
\end{table}
\begin{figure}[t]
    \centering
    \begin{subfigure}{0.25\textwidth}
      \includegraphics[width=\linewidth]{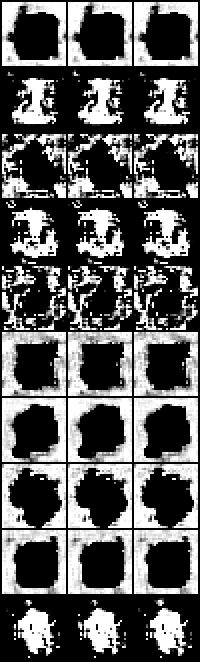}
      \caption{}
    \end{subfigure}
    \hfill
    \begin{subfigure}{0.25\textwidth}
      \includegraphics[width=\linewidth]{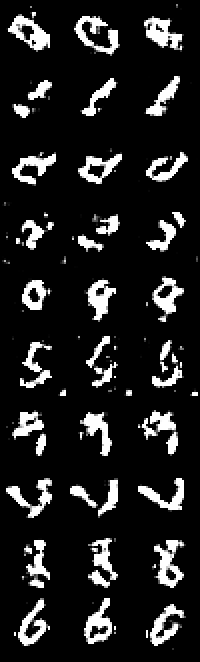}
      \caption{}
    \end{subfigure}
    \hfill
    \begin{subfigure}{0.25\textwidth}
    \includegraphics[width=\linewidth]{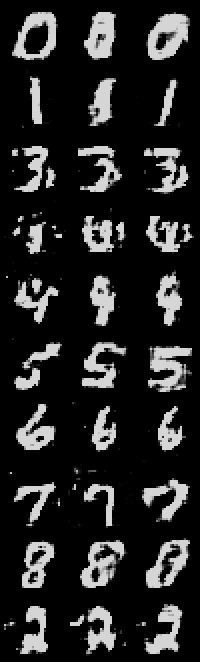}
      \caption{}
    \end{subfigure}
    \caption{Shape masks obtained after (a) addition of Gaussian random noise, (b) application of random rotation and (c) decrease of the mask opacity (i.e., lowering its transparency to 75\%).}
    \label{fig:shape_exp}
\end{figure}

As an additional contribution, we conduct a thorough study on the composition mechanism, to gain a better understanding of how the mechanisms influence the classification accuracy of an invariant classifier (i.e., a classifier that predicts whether an image is CGN-generated or real). Due to the limited time at our disposal, we focused on the mechanism that we deem most important in the decision-making of such a classifier, namely the \textit{shape}. To evaluate the effects of the shape IM we trained several (original) CGN models on the double-colored MNIST dataset; we tuned the resulting shape masks prior to the counterfactual image generation (governed by the composition mechanism of Equation \ref{eq:composition}) and used the generated images to train an invariant classifier. More specifically, we experimented with (i) the addition of Gaussian noise in the shape mask, (ii) random rotation of the mask, and (iii) multiplying the mask $\mathbf{m}$ in the composition mechanism (Eq. \ref{eq:composition}) with a factor smaller than 1 (or in other words, lowering the opacity of the shape mask). A transparency of 75\% (hence a weight factor of $0.75$) was experimentally found to be most beneficial for the accuracy of the classifier. 

The influence of the three transformations on the invariant classifier is quantified -- in terms of accuracy -- in Table \ref{tab:shape_exp_results}; sample shape masks generated from each transformation are displayed in Figure \ref{fig:shape_exp}. It is apparent from the test accuracy values that Gaussian noise and random rotations do not lead to any remarkable performance of the classifier but, contrarily, degrade its accuracy to values below 15\%. This is most likely the result of overfitting on the training set, as supported by the \textit{train} accuracy values. On the other hand, lowering the opacity of the mask substantially boosts the test accuracy, improving the previous results by a factor of $4\times$ (circa). It is noteworthy that the masks obtained using the transparency adjustment are more akin to those achieved using regular CGNs (see Figure \ref{fig:shape_exp}). The other transformations, instead, result in mask shapes that are particularly different. As such, they can potentially be used to make classifiers more robust when mixed with regular data during training. Because this is an extensive topic, we believe it warrants further research.



\section{Discussion and conclusion}
\label{sec:conclusion}




With the prevalence of heavily parameterized architectures such as BigGANs, and with the advent of limited-access models like the trending DALL·E 2, source-free compression becomes a growing necessity. In this paper we explored the possibility to obtain a lightweight version of the CGN network, assuming that we do not have access to the training process of the model. More specifically, we treat the backbone of each independent mechanism (shape, texture and background) as a black-box, then use KD to transfer the knowledge of the pre-trained cumbersome networks to simple TinyGANs.

SKDCGN achieves a remarkable compression of the overall network: it models the shape mechanism -- initially controlled by a BigGAN (55.9M parameters) and a U2-Net (44M parameters) -- using a single TinyGAN (6.4M parameters); similarly, it replaces the BigGANs responsible for the texture and background IMs with TinyGANs, and discards the forth BigGAN of the original CGN network that provides primary training supervision via reconstruction loss. This translates into four BigGANs and one U2-net (55.9M$\times$4 + 44M parameters, totalling 267.6M) being replaced with three simple TinyGANs (6.4M parameters each, meaning 19.2M parameters in total).

Despite the significant compression, we demonstrate the ability of our model to generate counterfactual images on ImageNet-1k and double-colored MNIST datasets (see Figure \ref{fig:im-results_t_b}). When trained on the latter, SKDCGN's network capacity was proven to be sufficient to model the simple data distribution. If trained on the former, the proposed method exhibited remarkable ability in mimicking the original shape and background generations, while the texture mechanism suffered more from the reduction of size. This finding reveals great potential for future works that would attempt to tune the distillation (and hence enhance the synthesis) of the texture images, for instance by including data augmentation in the training procedure.

Given the obtained results, we attemptedly limit the presence of noisy and ambiguous artifacts by minimizing the entropy between the teacher and student networks. We introduce a new measure in the knowledge distillation loss, i.e. KL divergence, which we find to enhance the visual quality results of some IMs for both Imagenet-1k and MNIST. 

Finally, we conduct a study on the composition mechanism to gain a better understanding of how the \textit{shape} IM influences the classification accuracy of an invariant classifier. Though other adjustments were tested, giving a lower weight to the shape mask $\mathbf{m}$ seemingly boosts the classifier performance.

\section{Future work}
To conclude, the experimental findings of SKDCGN prove that, upon the usage of Knowledge Distillation, one can transfer the capacity/ability of a cumbersome network to a lower-capacity model while still maintaining competitive performances. Although this paper unveils its potential, SKDCGN requires further research that we encourage other researchers to undertake. In addition to the suggestions offered throughout the sections, possible avenues of research include and are not limited to: improving the image generation process by using higher-order activation functions, since the utilized datasets consist of rich image data; improving the teacher-student architecture by introducing additional loss functions; using a learnable, neural network-based composition function instead of an analytical expression.

\section*{Acknowledgments} 
We would like to express our sincere gratitude to Prof. dr. Efstratios Gavves and Prof. Wilker Aziz for effectively organizing the \textit{Deep Learning II} course at the University of Amsterdam, which is the main reason this paper exists. We are thankful to our supervisor, Christos Athanasiadis, for his precious guidance throughout the project. Finally, we also thank the former Program Director of the MSc. Artificial Intelligence, Prof. dr. Cees G.M. Snoek, and the current Program Manager, Prof. dr. Evangelos Kanoulas, for effectively conducting the Master's program in Artificial Intelligence at the University of Amsterdam.


\clearpage

\appendix
\section*{Appendix}

\section{Architecture details of the different models}
This section contains the architectural details of the different model used in the proposed method. It brushes up the theory of the papers whereon we base our work (i.e. the CGN network \cite{DBLP:journals/corr/abs-2101-06046}, Sec. \ref{app:cgn-architecture} and the TinyGAN model \cite{DBLP:journals/corr/abs-2009-13829}, Sec. \ref{app:tinygan-architecture}) and also presents the baseline model (Sec. \ref{app:baseline-model}).
\subsection{Original CGN architecture}
\label{app:cgn-architecture}
This section contains a diagram of the original CGN architecture, as presented in \cite{DBLP:journals/corr/abs-2101-06046}. 

\begin{figure}[h]
  \centering
  \includegraphics[width=0.7\linewidth]{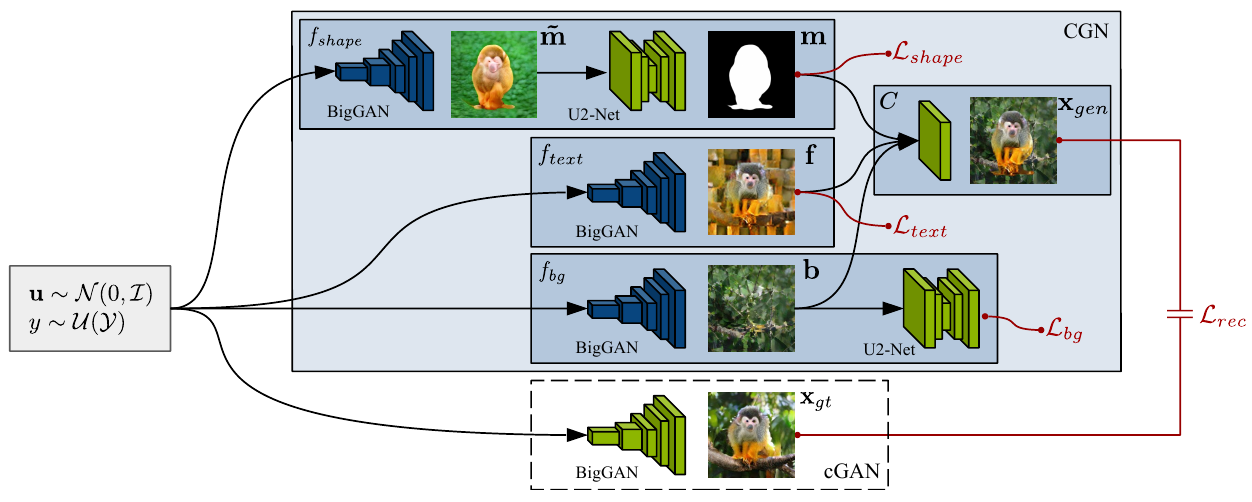}
  \caption{CGN architecture diagram. Retrieved from \cite{DBLP:journals/corr/abs-2101-06046}.}
  \label{fig:cgn-diagram}
\end{figure}

Figure \ref{fig:cgn-diagram} illustrates the CGN architecture. The network is split into four mechanisms, the shape mechanism $f_{shape}$, the texture mechanism $f_{text}$, the background mechanism $f_{bg}$, and the composer $C$. Components with trainable parameters are blue, components with fixed parameters are green. The primary supervision is provided by an unconstrained conditional GAN (cGAN) via the reconstruction loss $\mathcal{L}_{rec}$. The cGAN is only used for training, as indicated by the dotted lines. Each mechanism takes as input the noise vector $\mathbf{u}$ (sampled from a spherical Gaussian) and the label $y$ (drawn uniformly from the set of possible labels $\mathcal{Y}$) and minimizes its respective loss ($\mathcal{L}_{shape}$, $\mathcal{L}_{text}$, and $\mathcal{L}_{bg}$). To generate a set of counterfactual images, we sample $\mathbf{u}$ and then independently sample $y$ for each mechanism.

\subsection{TinyGAN architecture}
\label{app:tinygan-architecture}

\begin{figure}[t]
    \centering
    \begin{subfigure}{0.3\textwidth}
      \centering
      \includegraphics[width=1\linewidth]{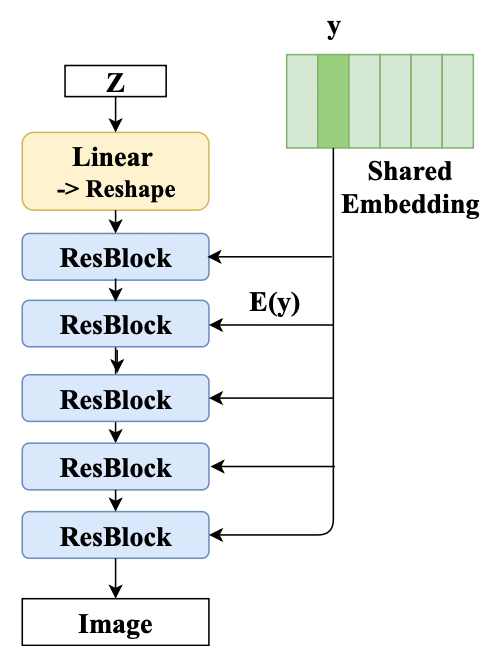}
      \caption{Student Generator $G$ \cite{DBLP:journals/corr/abs-2009-13829}
      }
      \label{fig:student generator}
    \end{subfigure}
    \begin{subfigure}{0.35\textwidth}
      \centering
      \includegraphics[width=1\linewidth]{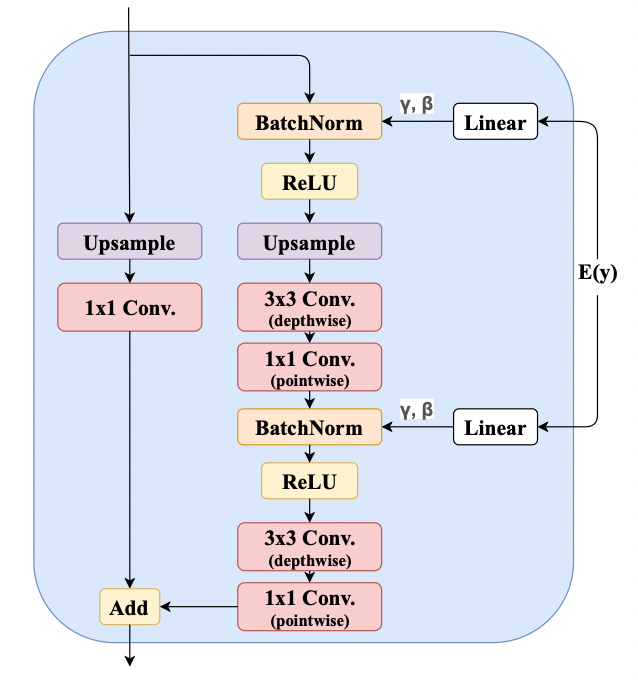}
      \caption{A Residual Block in $G$ \cite{DBLP:journals/corr/abs-2009-13829}
      }
      \label{fig:residual block}
    \end{subfigure}
    \caption{Architecture of the TinyGAN (student) generator}
    \label{fig:tinygan-generator}
\end{figure}

This section provides an brief overview of the TinyGAN architecture. For more details, refer to \cite{DBLP:journals/corr/abs-2009-13829}.

\paragraph{Generator.} As shown in Figure \ref{fig:tinygan-generator}, TinyGAN comprises a ResNet \cite{resnet}-based generator with class-conditional BatchNorm \cite{batchnorm1} \cite{batchnorm2}. To keep a tight computation budget, it does not adopt attention-based \cite{self-attention} or progressive-growing mechanisms \cite{progressing-growing}. To substantially reduce the model size compared to BigGAN, it:
\begin{itemize}
    \item Relies on using fewer channels;
    \item Replaces standard convolution by depthwise separable convolution;
    \item Adopts a simpler way to introduce class conditions.
\end{itemize}
Overall, TinyGAN's generator has 16$\times$ less parameters than BigGAN's generator.

\vspace{-0.5em}

\paragraph{Discriminator.} Following \cite{ref-discr-1}
\cite{DBLP:journals/corr/abs-1802-05957}, \cite{DBLP:journals/corr/abs-2009-13829} opt for spectral normalized discriminator and introduce the class condition via projection. But instead of utilizing complicated residual blocks, they simply stack multiple convolutional layers with stride as used in DCGAN \cite{dcgan}, which greatly reduces the number of parameters. 

Overall, TinyGAN's discriminator has 10$\times$ less parameters than BigGAN's discriminator.

\subsection{Baseline model}
\label{app:baseline-model}
The baseline is a standard CGN architecture whose BigGANs have been replaced with TinyGANs. Due to the need of a pre-trained model that (i) supervises the CGN training using a reconstruction loss and (ii) serves as the initialization of the IM GANs, a TinyGAN was trained from scratch using the KD strategy described in \cite{DBLP:journals/corr/abs-2009-13829}. Section \ref{app:baseline-details} dives into the details of the training procedure, then presents qualitative results of both the newly-trained TinyGAN and of baseline model.

\begin{figure}[t!]
  \begin{subfigure}{\textwidth}
  \centering
  \hspace{6mm} \textit{ImageNet-1k} \hspace{36mm} \textit{Double-colored MNIST}\\
      \includegraphics[width=0.48\linewidth]{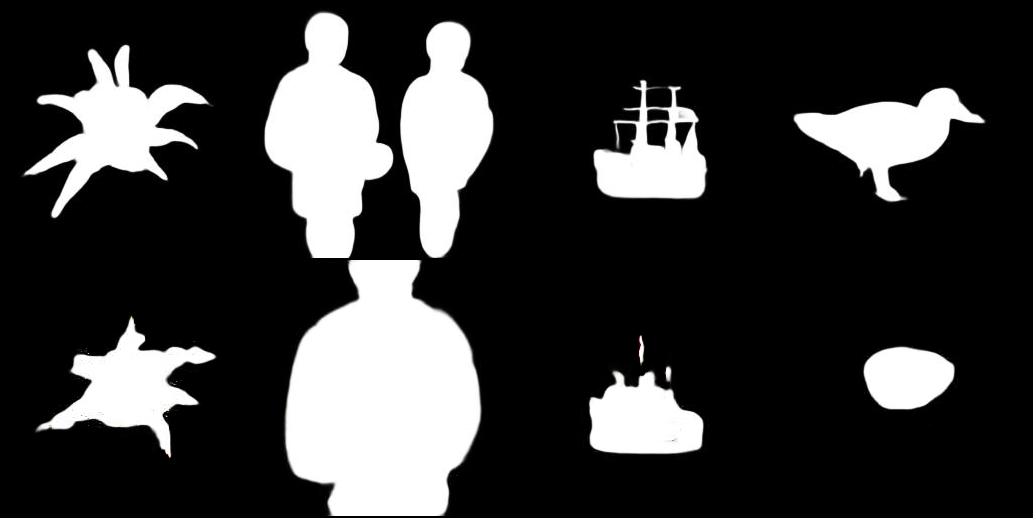}
      \hfill
      \includegraphics[width=0.48\linewidth]{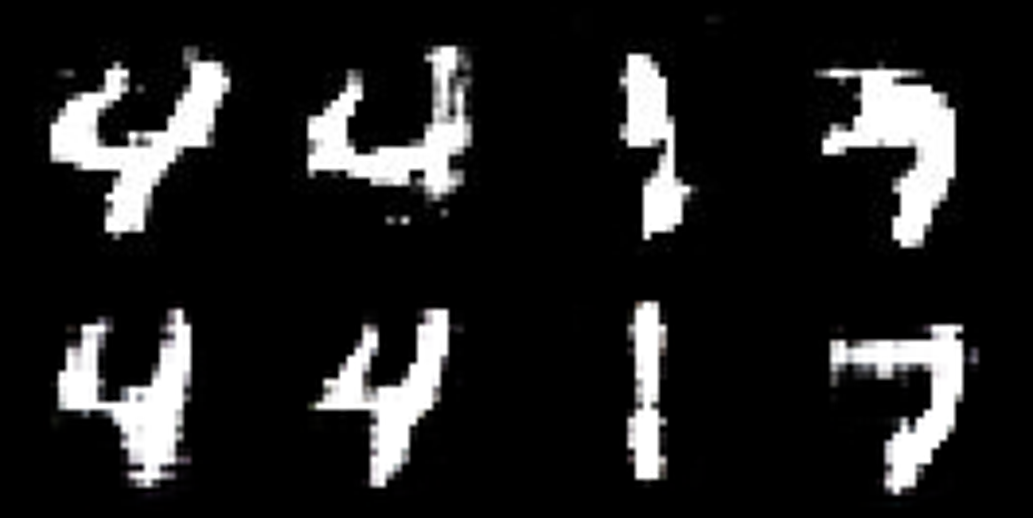}
      \caption{\textit{Shape} mechanism.}
      \label{fig:shape_results}
    \end{subfigure}
    \\
  \begin{subfigure}{\textwidth}
  \centering
  \hspace{6mm} \textit{ImageNet-1k} \hspace{36mm} \textit{Double-colored MNIST}\\
      \includegraphics[width=0.48\linewidth]{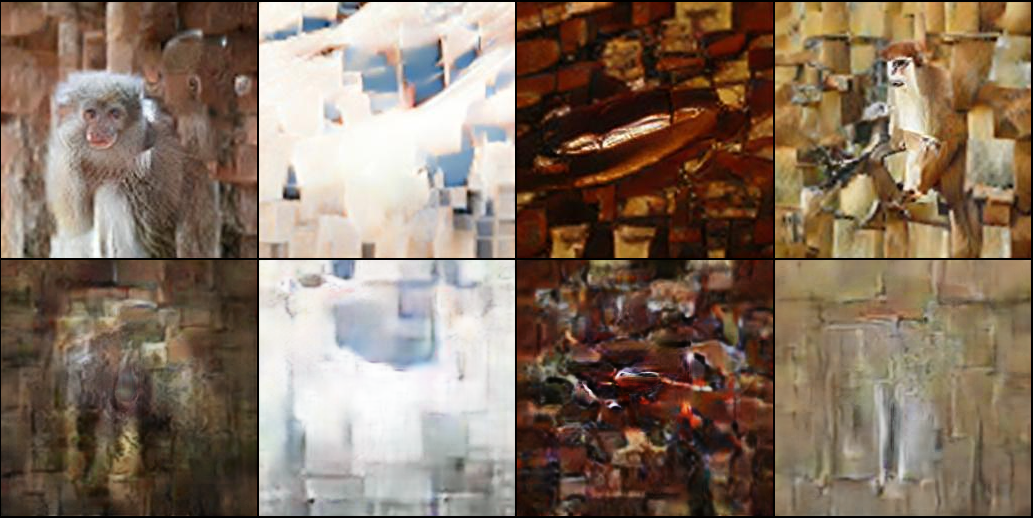}
      \hfill
      \includegraphics[width=0.48\linewidth]{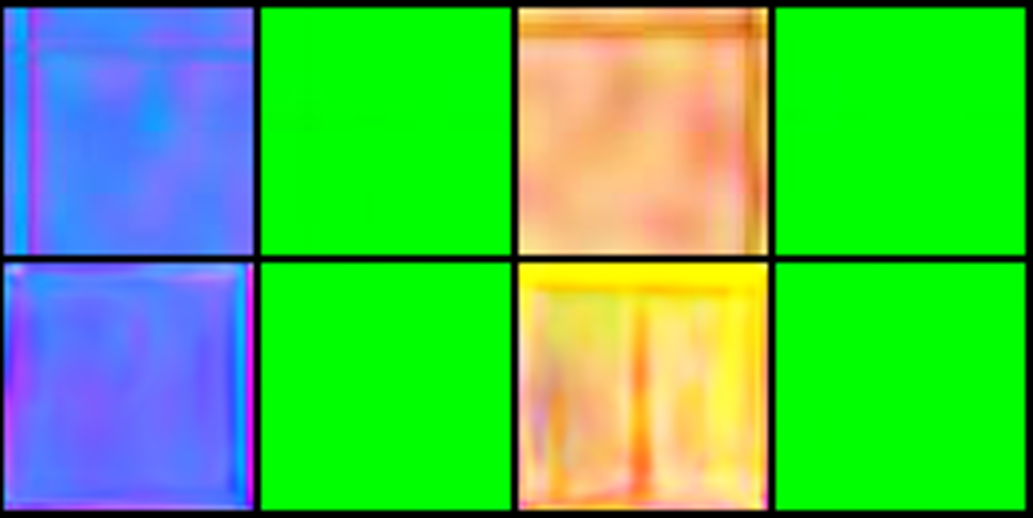}
      \caption{\textit{Texture} mechanism.}
      \label{fig:fg_results}
  \end{subfigure}
  \\
    \begin{subfigure}{\textwidth}
    \centering
      \hspace{6mm} \textit{ImageNet-1k} \hspace{36mm} \textit{Double-colored MNIST}\\
      \includegraphics[width=0.48\linewidth]{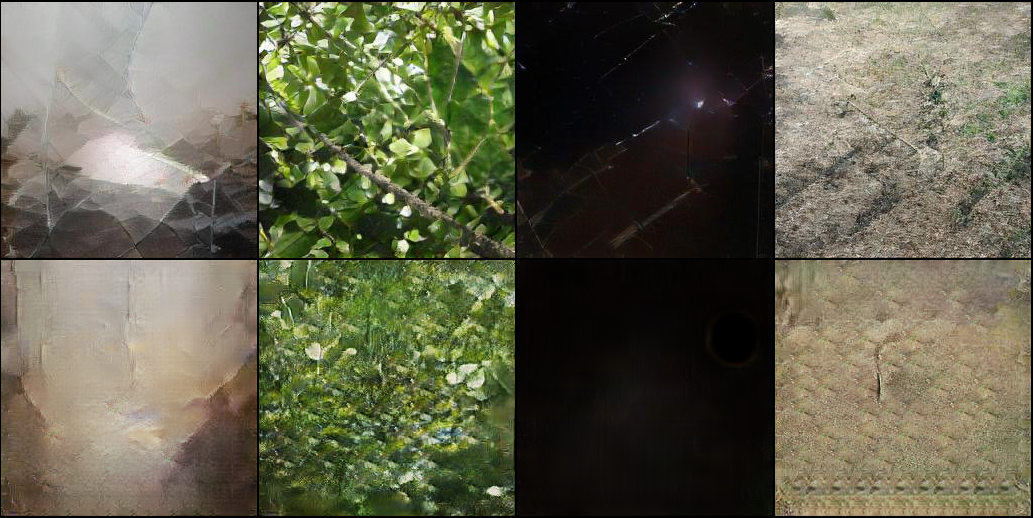}
      \hfill
      \includegraphics[width=0.48\linewidth]{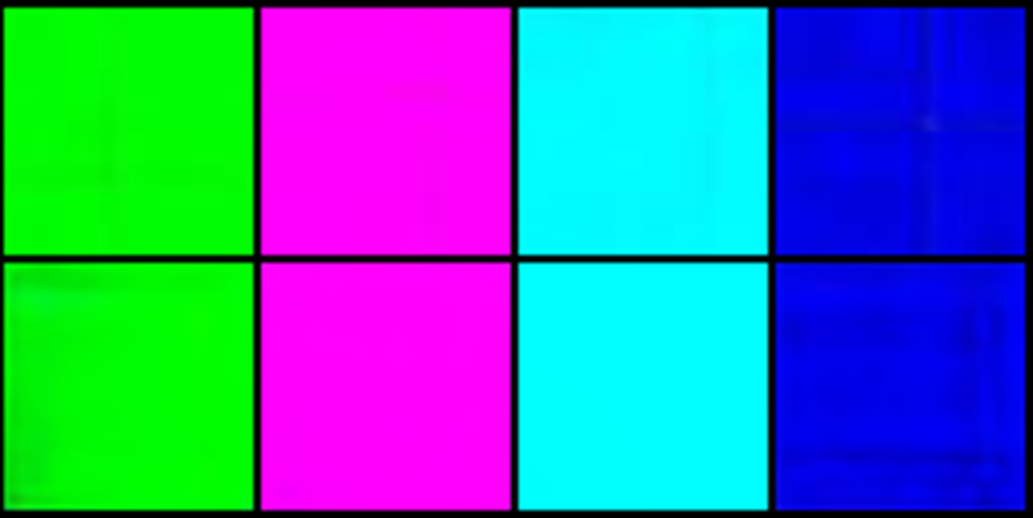}
      \caption{\textit{Background} mechanism.}
      \label{fig:bg_results}
  \end{subfigure}
  \caption{A comparison of images generated by the CGN backbones and those generated by the corresponding SKDCGN's TinyGAN (given the same input), for each independent mechanism. We train on both ImageNet-1k (left images) and double-colored MNIST datasets (right images).}
  \label{fig:mnist_ims}
\end{figure}

\section{Additional results of SKDCGN's IMs}
This section expands Section 4.3 of the main paper and contains more results obtained from each SKDCGN's IM, using both ImageNet-1k and double-colored MNIST datasets. More specifically, we compare the output of each CGN backbone with that of the corresponding SKDCGN's TinyGAN, given the same input. Please refer to Figure \ref{fig:mnist_ims}.

\section{Baseline Model}
\label{app:baseline-details}
The baseline model is a modified version of the original CGN architecture, where each BigGAN has been replaced by the generator model of a TinyGAN. Training this baseline using the procedure described by \cite{DBLP:journals/corr/abs-2009-13829}, omitting KD, allows for rigorous comparisons that emphasize the effectiveness of the knowledge distillation process. In this section we provide training details, and collect sample outputs of the trained model.

\begin{figure}[t!]
  \begin{subfigure}{\textwidth}
  \includegraphics[width=\linewidth]{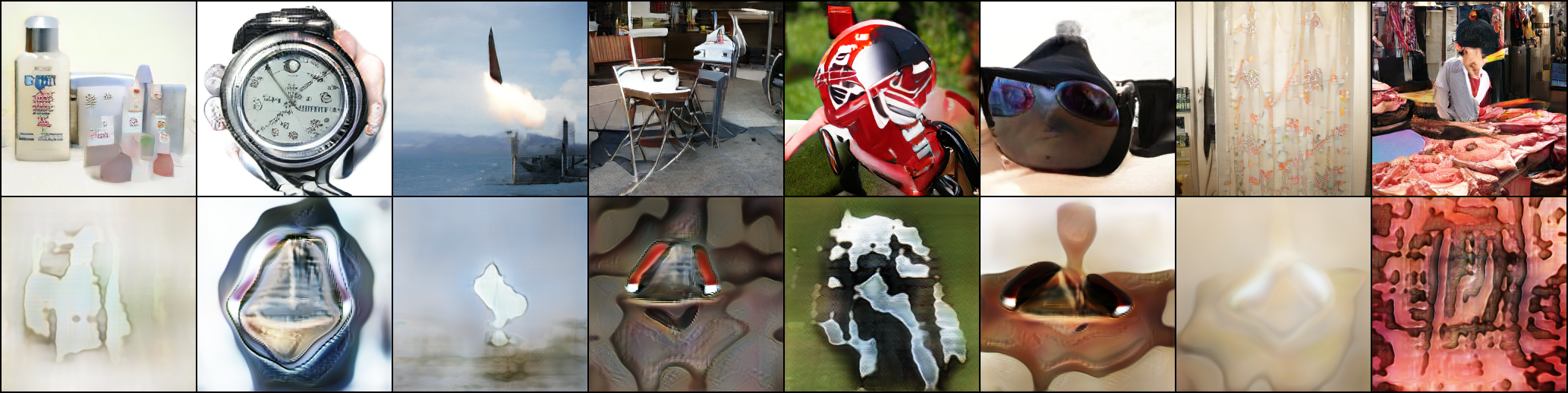}
  \caption{A comparison of images generated by BigGAN and the TinyGAN. Images in top row are produced by BigGAN, while those in bottom row are by SKDCGN given the same input after $1^{st}$ epoch.}
  \label{fig:tinygan_results_1}
  \end{subfigure}
  \begin{subfigure}{\textwidth}
  \includegraphics[width=\linewidth]{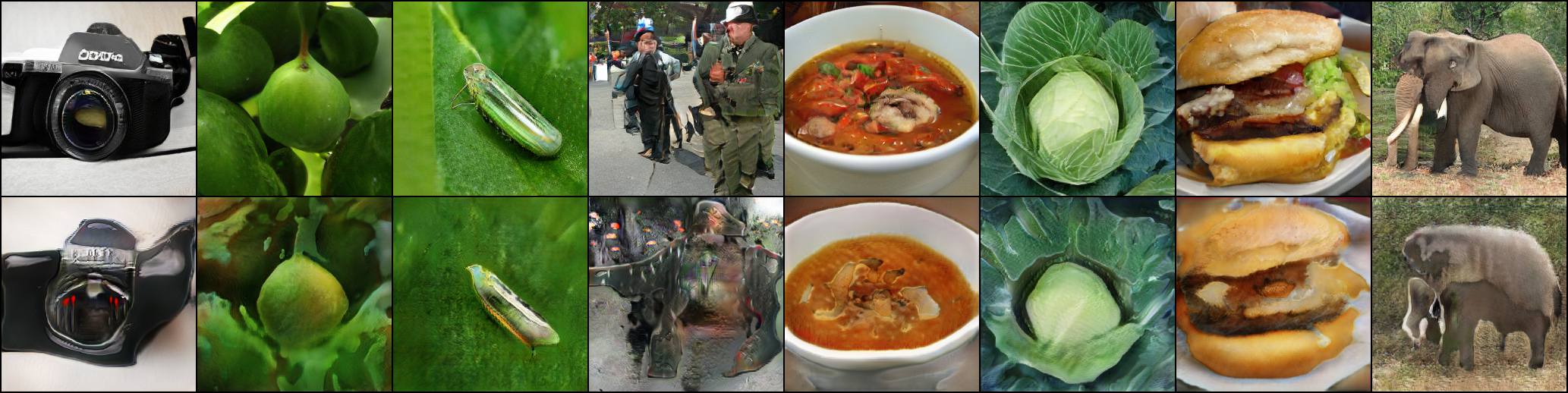}
  \caption{A comparison of images generated by BigGAN and the TinyGAN. Images in top row are produced by BigGAN, while those in bottom row are by SKDCGN given the same input after $18^{th}$ epoch.}
  \label{fig:tinygan_results_18}
  \end{subfigure}
  \caption{A comparison of images generated by BigGAN and the TinyGAN. Images in top row are produced by BigGAN, while those in bottom row are by SKDCGN given the same input}
  \label{tinygan_results}
\end{figure}

\subsection{Training Details}
\label{app:baseline-training}
The training procedure of a CGN requires a pre-trained GAN to provide primary supervision via the reconstruction loss. However, the original TinyGAN was only trained on only animal classes, hence the publicly-available model could not be used for our baseline. In order to consistently use the same dataset for all the experiments, we re-trained a TinyGAN from scratch (as described in \cite{DBLP:journals/corr/abs-2009-13829}) on all classes of ImageNet-1k. The images generated by TinyGAN are visualized in Appendix \ref{app:pretrained-tinygan-gen-outputs}. The images generated for each Independent Mechanism using our baseline model can be seen in \ref{app:baseline-outputs}. Apart from this, we additionally generated the counterfactuals using the baseline model which are shown in Appendix \ref{app:baseline-counterfactuals}.

\begin{figure}[ht!]
\centering
\begin{tabular}{lllll}
$\Tilde{m}$ &
\includegraphics[width=.18\linewidth]{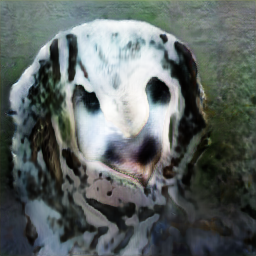}
\hspace{-0.5em}
\includegraphics[width=.18\linewidth]{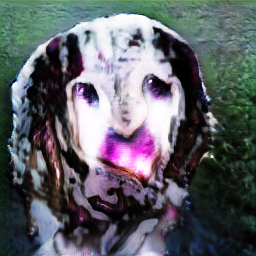}
\hspace{-0.5em}
\includegraphics[width=.18\linewidth]{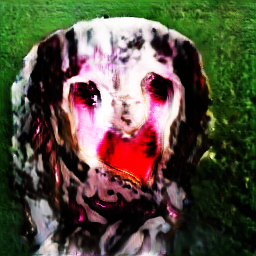}
\hspace{-0.5em}
\includegraphics[width=.18\linewidth]{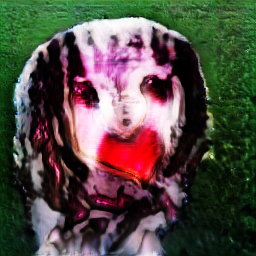}
\hspace{-0.5em}
\includegraphics[width=.18\linewidth]{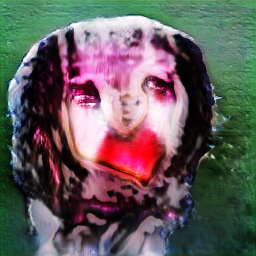}
\vspace{-0.31em}\\
\vspace{-0.34em}
$m$ &
\includegraphics[width=.18\linewidth]{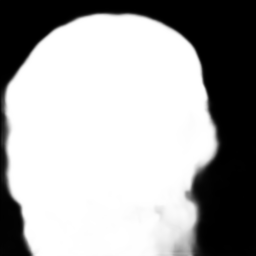}
\hspace{-0.49em}
\includegraphics[width=.18\linewidth]{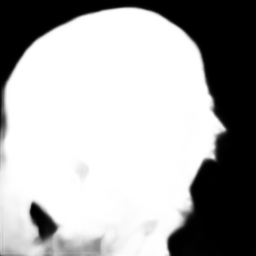} 
\hspace{-0.49em}
\includegraphics[width=.18\linewidth]{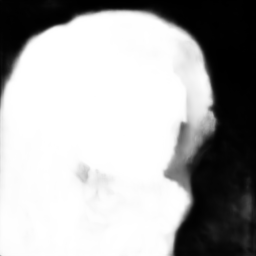} 
\hspace{-0.48em}
\includegraphics[width=.18\linewidth]{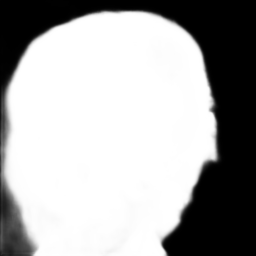} 
\hspace{-0.5em}
\includegraphics[width=.18\linewidth]{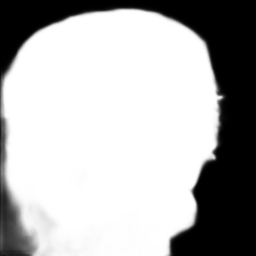}\\
\vspace{-0.33em}
$f$ &
\includegraphics[width=.18\linewidth]{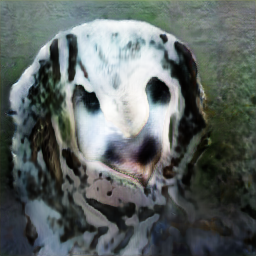}
\hspace{-0.49em}
\includegraphics[width=.18\linewidth]{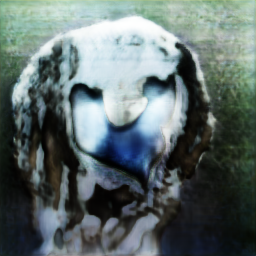} 
\hspace{-0.49em}
\includegraphics[width=.18\linewidth]{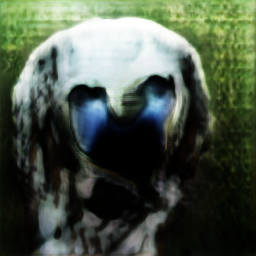} 
\hspace{-0.48em}
\includegraphics[width=.18\linewidth]{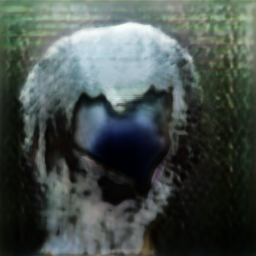} 
\hspace{-0.5em}
\includegraphics[width=.18\linewidth]{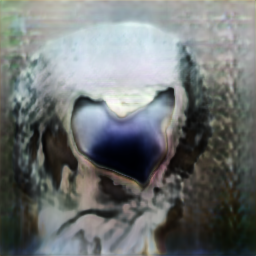}\\
\vspace{-0.33em}
$b$ &
\includegraphics[width=.18\linewidth]{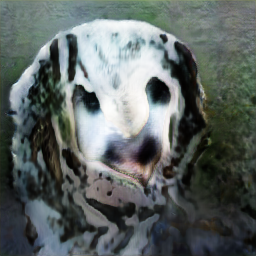}
\hspace{-0.49em}
\includegraphics[width=.18\linewidth]{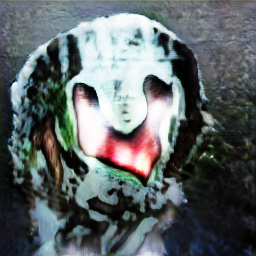} 
\hspace{-0.49em}
\includegraphics[width=.18\linewidth]{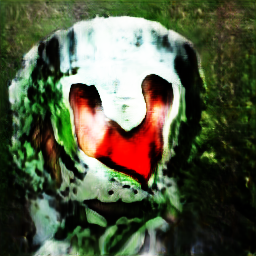} 
\hspace{-0.48em}
\includegraphics[width=.18\linewidth]{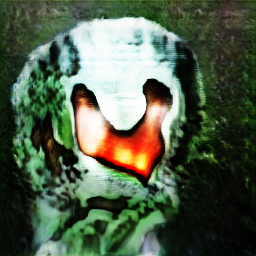} 
\hspace{-0.5em}
\includegraphics[width=.18\linewidth]{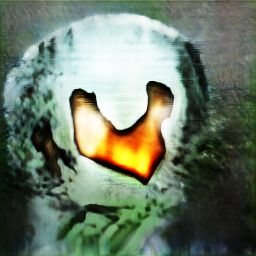}\\
\vspace{-0.41em}
$x_{gen}$ &
\includegraphics[width=.18\linewidth]{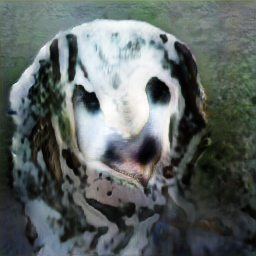}
\hspace{-0.49em}
\includegraphics[width=.18\linewidth]{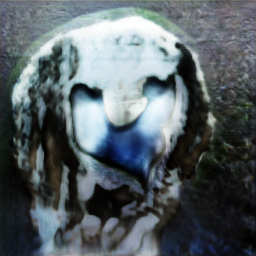} 
\hspace{-0.49em}
\includegraphics[width=.18\linewidth]{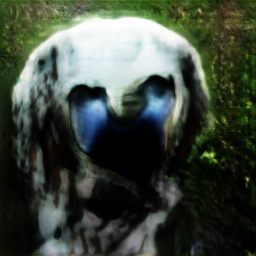} 
\hspace{-0.48em}
\includegraphics[width=.18\linewidth]{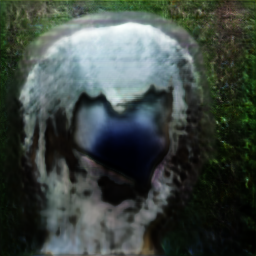} 
\hspace{-0.5em}
\includegraphics[width=.18\linewidth]{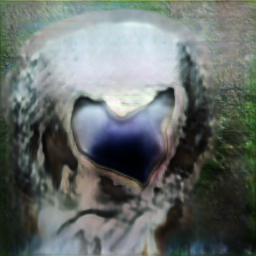}
\end{tabular}
\caption{Individual IM Outputs after training for baseline. From top to bottom: $m$, $\Tilde{m}$, $f$, $b$, $x_{gen}$. From left to right: at the start of training, after epoch $300k{th}$, epoch $600k^{th}$, epoch $900k^{th}$, and epoch $1.2million^{th}$}
\label{fig:IMs_baseline_2}
\end{figure}

\subsubsection{Generated outputs of TinyGAN trained on ImageNet-1k}
\label{app:pretrained-tinygan-gen-outputs}
A TinyGAN was trained using all 1000 classes of the ImageNet-1k dataset. Training details are provided by \cite{DBLP:journals/corr/abs-2009-13829}. Although the original paper trains the model for 1.2 million epochs, we are forced to restrict the amount of iterations due to computational constraints. After distilling the knowledge of a BigGAN for 18 epochs, our TinyGAN generates reasonable images, as seen in Figure \ref{fig:tinygan_results_18}. To compare the image generation we have also presented images generated after the first epoch as well \ref{fig:tinygan_results_1}. It can be observed that if we further train the model, it could produce images better in quality. Note that animal classes are better captured by the model: this is inline with the findings of \cite{DBLP:journals/corr/abs-2009-13829}.

\begin{figure}[ht!]
\centering
\begin{tabular}{lllll}

\includegraphics[width=.18\linewidth]{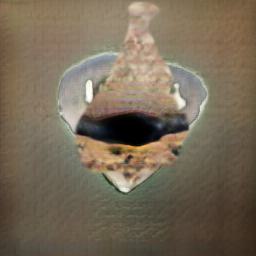}
\hspace{-0.49em}
\includegraphics[width=.18\linewidth]{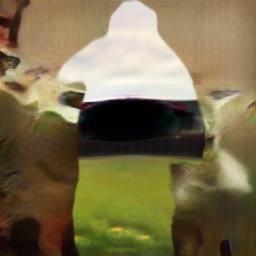}
\hspace{-0.49em}
\includegraphics[width=.18\linewidth]{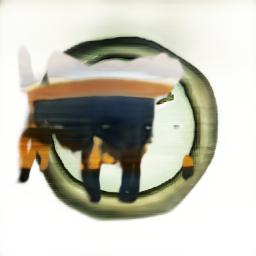}
\hspace{-0.49em}
\includegraphics[width=.18\linewidth]{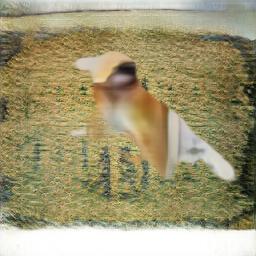}
\hspace{-0.5em}
\includegraphics[width=.18\linewidth]{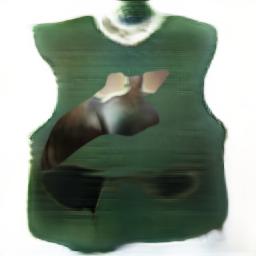}
\vspace{-0.31em}\\
\vspace{-0.33em}

\includegraphics[width=.18\linewidth]{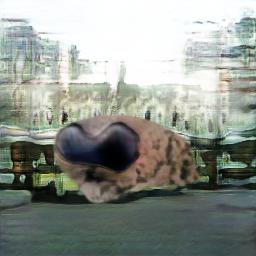}
\hspace{-0.49em}
\includegraphics[width=.18\linewidth]{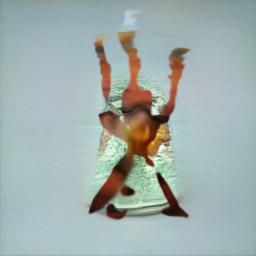} 
\hspace{-0.49em}
\includegraphics[width=.18\linewidth]{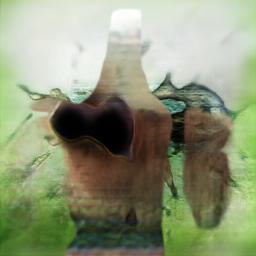} 
\hspace{-0.49em}
\includegraphics[width=.18\linewidth]{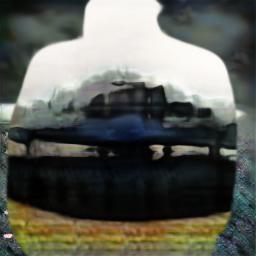} 
\hspace{-0.5em}
\includegraphics[width=.18\linewidth]{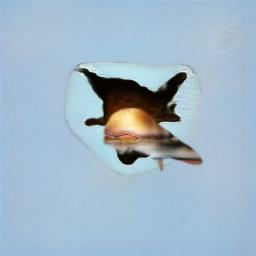}\\
\vspace{-0.32em}

\includegraphics[width=.18\linewidth]{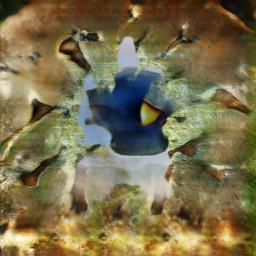}
\hspace{-0.49em}
\includegraphics[width=.18\linewidth]{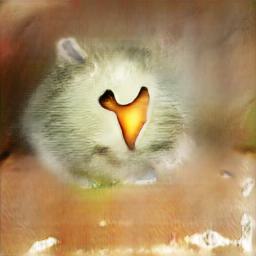} 
\hspace{-0.49em}
\includegraphics[width=.18\linewidth]{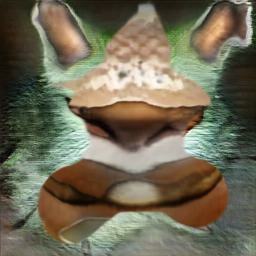} 
\hspace{-0.49em}
\includegraphics[width=.18\linewidth]{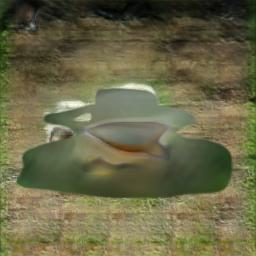} 
\hspace{-0.5em}
\includegraphics[width=.18\linewidth]{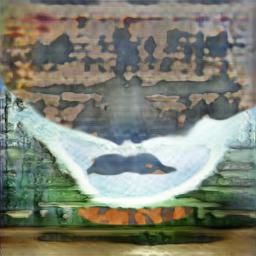}\\
\vspace{-0.32em}

\includegraphics[width=.18\linewidth]{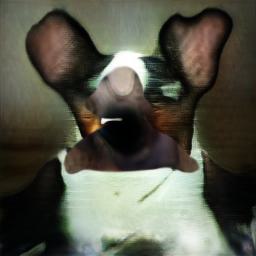}
\hspace{-0.49em}
\includegraphics[width=.18\linewidth]{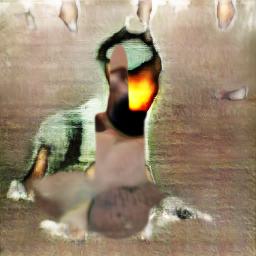} 
\hspace{-0.49em}
\includegraphics[width=.18\linewidth]{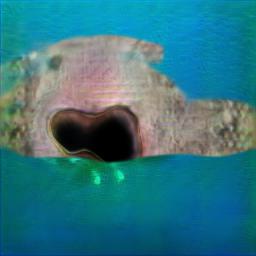} 
\hspace{-0.49em}
\includegraphics[width=.18\linewidth]{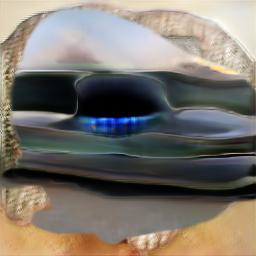} 
\hspace{-0.5em}
\includegraphics[width=.18\linewidth]{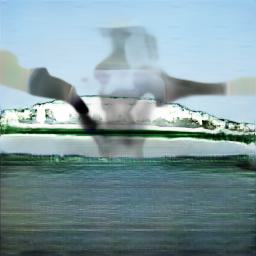}\\

\vspace{-0.4em}
\includegraphics[width=.18\linewidth]{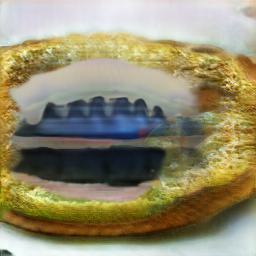}
\hspace{-0.49em}
\includegraphics[width=.18\linewidth]{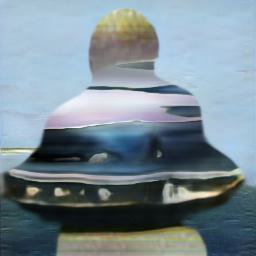} 
\hspace{-0.49em}
\includegraphics[width=.18\linewidth]{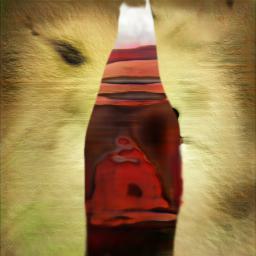} 
\hspace{-0.49em}
\includegraphics[width=.18\linewidth]{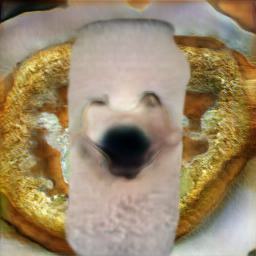} 
\hspace{-0.5em}
\includegraphics[width=.18\linewidth]{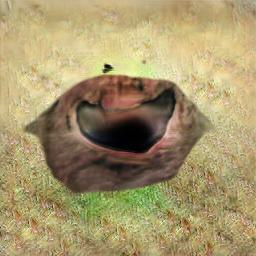}
\end{tabular}
\caption{Counterfactuals generated by baseline on test data for ImageNet-1k}
\label{fig:counterfactuals_baseline}
\end{figure}

\subsubsection{Generated outputs of the baseline trained on ImageNet-1k}
\label{app:baseline-outputs}
Figure \ref{fig:IMs_baseline_2} illustrates the individual outputs of each IMs at the start of training, after epoch 300k$^{\text{th}}$, epoch 600k$^{\text{th}}$, epoch 900k$^{\text{th}}$, and epoch 1.2M$^{\text{th}}$ (from left to right). In each figure, we show from top to bottom: pre-masks $\Tilde{m}$, masks $m$, texture $f$, background $b$, and composite images $x_{gen}$.

\subsubsection{Generated Counterfactual Images of Baseline trained on ImageNet-1k}
\label{app:baseline-counterfactuals}
Finally, we show  counterfactual images generated by the baseline model in Figure \ref{fig:counterfactuals_baseline}.

\section{Improving the SKDCGN process} \label{sec:improve_skdcgn}
\begin{figure}[ht]
  \centering
  \includegraphics[width=\linewidth]{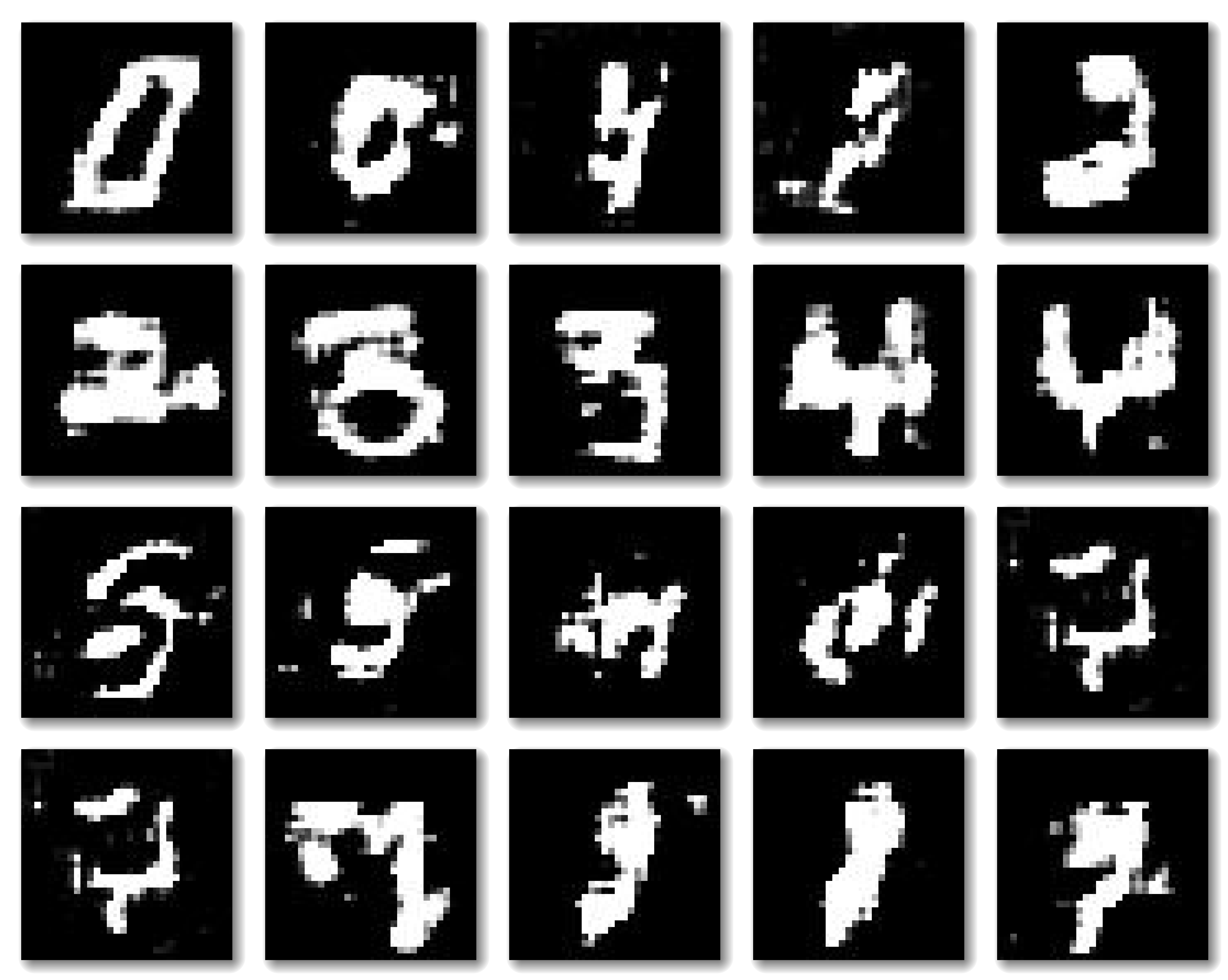}
  \caption{Noisy outputs generated by the CGN when we made use of pretrained weights given by the authors. }
  \label{fig:mnist_cgn_noisy}
\end{figure}

As mentioned in Section 4.4 of the main paper, we observed that the outputs from CGN are noisy in nature. Fig \ref{fig:mnist_cgn_noisy} evidently illustrates how noisy the MNIST digits are. However in this section we try to improve our architecture by several methods. 

In the direction towards improving the images that are being generated by our architecture, we strongly believe the room of improvement lies in these components:
\begin{itemize}
    \item Improving the quality of images that are being generated by the GAN network in our architecture. Usually loss functions like VGG based perception loss, L1 reconstruction loss are added. 
    \item Improving the existing knowledge distillation framework such that the student learns better from the teacher's guidance by adding new loss functions to the Knowledge Distillation task. 
\end{itemize}

\begin{figure}[ht!]
\centering
  \begin{subfigure}{\textwidth}
      \includegraphics[width=\linewidth]{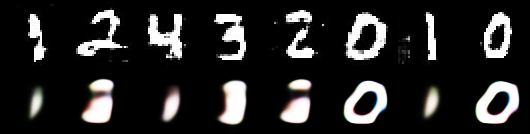}
      \caption{A comparison of images generated by the CGN \textbf{shape} backbone (\textit{top} row) and those generated by the corresponding SKDCGN given the same input (\textit{bottom} row) after 2 epochs on test data.}
      \label{fig:mnist_mask_bce2}
  \end{subfigure}
  \\
  \begin{subfigure}{\textwidth}
      \includegraphics[width=\linewidth]{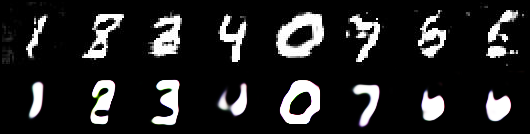}
      \caption{A comparison of images generated by the CGN \textbf{texture} backbone (\textit{top} row) and those generated by the corresponding SKDCGN given the same input (\textit{bottom} row) after 10 epochs on test data.}
      \label{fig:mnist_mask_bce10}
  \end{subfigure}
  \\
    \begin{subfigure}{\textwidth}
      \includegraphics[width=\linewidth]{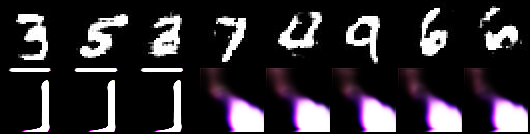}
      \caption{A comparison of images generated by the \textbf{background} backbone (\textit{top} row) and those generated by the corresponding SKDCGN given the same input (\textit{bottom} row) after 30 epochs on test data.}
      \label{fig:mnist__mask_bce30}
  \end{subfigure}
  \caption{A comparison of images generated by the CGN backbones and those generated by the corresponding SKDCGN (given the same input) mask IM with cross entropy loss}
  \label{fig:mnist_ims_1}
\end{figure}

To improve the quality of images, we observe that our architecture already has most of the loss functions integrated implicitly/explicitly. Hence, we add the Cross entropy loss for the generator and discriminator for the mask IM of the architecture and get the results as shown in \ref{fig:mnist_mask_bce2} for second epoch. We observe that digits like '0' are being reconstructed however for other digits the inputs look noisy in nature. By the end of 10th epoch for test set in Fig. \ref{fig:mnist_mask_bce10} we observe that the digits are being reconstructed. We continue with the training since we expected better results than what we have a;ready seen, however, contrary to our beliefs we observe artefacts by the end of 30th epoch as shown in Fig. \ref{fig:mnist__mask_bce30}.

\begin{figure}[ht!]
\centering
  \begin{subfigure}{0.9\textwidth}
      \includegraphics[width=\linewidth]{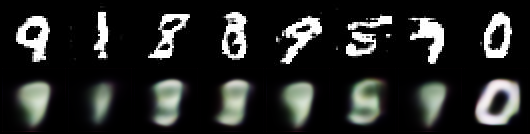}
      \caption{A comparison of images generated by the CGN \textbf{shape} backbone (\textit{top} row) and those generated by the corresponding SKDCGN given the same input (\textit{bottom} row) after 2 epochs on test data.}
      \label{fig:mnist_kl_layer2}
  \end{subfigure}
  \\
  \begin{subfigure}{0.9\textwidth}
      \includegraphics[width=\linewidth]{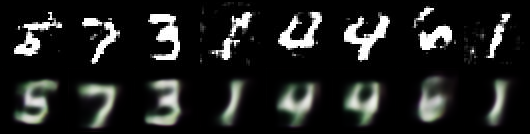}
      \caption{A comparison of images generated by the CGN \textbf{texture} backbone (\textit{top} row) and those generated by the corresponding SKDCGN given the same input (\textit{bottom} row) after 10 epochs on test data.}
      \label{fig:mnist_kl_layer10}
  \end{subfigure}
  \\
    \begin{subfigure}{0.9\textwidth}
      \includegraphics[width=\linewidth]{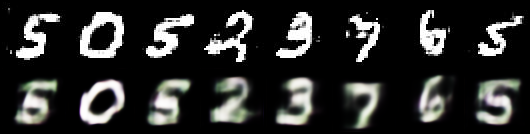}
      \caption{A comparison of images generated by the \textbf{background} backbone (\textit{top} row) and those generated by the corresponding SKDCGN given the same input (\textit{bottom} row) after 30 epochs on test data.}
      \label{fig:mnist_kl_layer30}
  \end{subfigure}
  \caption{A comparison of images generated by the CGN backbones and those generated by the corresponding SKDCGN (given the same input) mask IM with KL divergence multiplied with the activation of every layer instead of L1}
  \label{fig:mnist_kl_layer}
\end{figure}

\subsection{KL multiplied with layer instead of L1} \label{app:kl_instead_l1}
Since the image generation process already has most of the components to ensure that the reconstruction is in place, we tried to improve the Knowledge distillation between teacher and student network by integrating the KL divergence and multiply the loss with every layer of the network instead of L1 which is default. Possibly, because L1 reconstruction loss is explicitly needed that is to multiplied with the activation of every layer.  We observe the results as shown in Fig. \ref{fig:mnist_kl_layer}

\begin{figure}[ht!]
  \begin{subfigure}{\textwidth}
      \includegraphics[width=\linewidth]{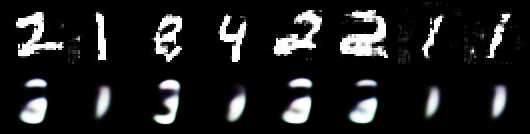}
      \caption{A comparison of images generated by the CGN \textbf{shape} backbone (\textit{top} row) and those generated by the corresponding SKDCGN given the same input (\textit{bottom} row) for mask IM after 2 epochs on test data.}
      \label{fig:mnist_mask_mse2}
  \end{subfigure}
  \\
  \begin{subfigure}{\textwidth}
      \includegraphics[width=\linewidth]{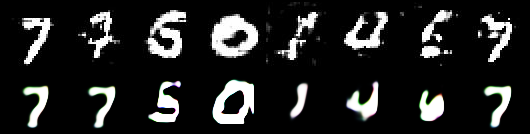}
      \caption{A comparison of images generated by the CGN \textbf{texture} backbone (\textit{top} row) and those generated by the corresponding SKDCGN given the same input (\textit{bottom} row) for mask IM after 10 epochs on test data.}
      \label{fig:mnist_mask_mse10}
  \end{subfigure}
  \\
    \begin{subfigure}{\textwidth}
      \includegraphics[width=\linewidth]{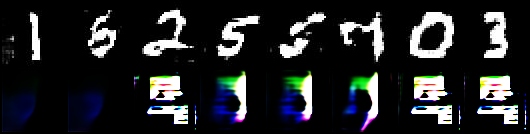}
      \caption{A comparison of images generated by the \textbf{background} backbone (\textit{top} row) and those generated by the corresponding SKDCGN given the same input (\textit{bottom} row) for mask IM after 30 epochs on test data.}
      \label{fig:mnist__mask_mse_30}
  \end{subfigure}
  \caption{A comparison of images generated by the CGN backbones and those generated by the corresponding SKDCGN (given the same input) mask IM with L2 multiplied with the activation of every layer instead of L1.}
  \label{fig:mnist_mse}
\end{figure}

\subsection{MSE instead of L1}  \label{app:mse_no_l1}
In addition, 
We also tried L2 loss instead of L1 loss but it lead to noisy outputs than previously generated and obtain results as shown in \ref{fig:mnist_mse}. Since, L2 assumes that the influence of noise is independent of the image's local characteristic the images are noisy in nature. 

\clearpage
%
%
\bibliographystyle{unsrt}
\bibliography{egbib}

\end{document}